%% file: 0_main.tex
\documentclass[referee,pdflatex]{sn-jnl}

\usepackage{subcaption}
\usepackage{graphicx}%
\usepackage{multirow}%
\usepackage{amsmath,amssymb,amsfonts}%
\usepackage{amsthm}%
\usepackage{mathrsfs}%
\usepackage[title]{appendix}%
\usepackage[dvipsnames, svgnames, x11names]{xcolor}%
\usepackage{textcomp}%
\usepackage{manyfoot}%
\usepackage{booktabs}%
\usepackage{algorithm}%
\usepackage{algorithmicx}%
\usepackage{algpseudocode}%
\usepackage{listings}%
\usepackage{longtable}
\usepackage{makecell}
\usepackage{lmodern}
\usepackage{mathrsfs}

\usepackage[square,numbers,sort&compress]{natbib}

\newcommand{\etal}{\textit{et al.}}
\newcommand{\revise}[1]{#1}

\begin{document}

\title[Fairness Survey]{Addressing Fairness Issues in Deep Learning-Based Medical Image Analysis: A Systematic Review}


\author[1,2]{\fnm{Zikang} \sur{Xu}}

\author[3]{\fnm{Jun} \sur{Li}}

\author[3]{\fnm{Qingsong} \sur{Yao}}

\author[1,2]{\fnm{Han} \sur{Li}}

\author[1,2]{\fnm{Mingyue} \sur{Zhao}}
\author*[1,2,3,4]{\fnm{S. Kevin} \sur{Zhou}}\email{skevinzhou@ustc.edu.cn}

\affil[1]{\orgdiv{School of Biomedical Engineering, Division of Life Sciences and Medicine}, \orgname{University of Science and Technology of China}, \orgaddress{\city{Hefei, Anhui}, \postcode{230026}, \country{P.R. China}}}

\affil[2]{\orgdiv{Center for Medical Imaging, Robotics, Analytic Computing \& Learning (MIRACLE), Suzhou Institute for Advanced Research}, \orgname{University of Science and Technology of China}, \orgaddress{\city{Suzhou, Jiangsu}, \postcode{215123}, \country{P. R. China}}}

\affil[3]{\orgdiv{Key Lab of Intelligent Information Processing of Chinese Academy of Sciences (CAS)}, \orgname{Institute of Computing Technology, CAS}, \orgaddress{\city{Beijing}, \postcode{100190}, \country{P. R. China}}}

\affil[4]{\orgdiv{Key Laboratory of Precision and Intelligent Chemistry}, \orgname{University of Science and Technology of China}, \orgaddress{\city{Hefei Anhui}, \postcode{230026}, \country{P. R. China}}}

\newpage

\abstract{
    Deep learning algorithms have demonstrated remarkable efficacy in various medical image analysis (MedIA) applications. However, recent research highlights a performance disparity in these algorithms when applied to specific subgroups, such as exhibiting poorer predictive performance in elderly females. Addressing this fairness issue has become a collaborative effort involving AI scientists and clinicians seeking to understand its origins and develop solutions for mitigation within MedIA. In this survey, we thoroughly examine the current advancements in addressing fairness issues in MedIA, focusing on methodological approaches. We introduce the basics of group fairness and subsequently categorize studies on fair MedIA into fairness evaluation and unfairness mitigation. Detailed methods employed in these studies are presented too. Our survey concludes with a discussion of existing challenges and opportunities in establishing a fair MedIA and healthcare system. By offering this comprehensive review, we aim to foster a shared understanding of fairness among AI researchers and clinicians, enhance the development of unfairness mitigation methods, and contribute to the creation of an equitable MedIA society.
    }




\maketitle

\section{Introduction}

The progressive advancement of artificial intelligence (AI) has garnered substantial attention and development in recent years, showcasing its efficacy in diverse practical applications such as autonomous driving, recommendation systems, and more. Notably, the data-centric approach inherent in AI methodologies has emerged as an indispensable asset in the domains of healthcare and medical image analysis.

Amidst the substantial research dedicated to refining the performance of machine learning (ML) or deep learning (DL) algorithms, a notable concern has surfaced among researchers. {Although the performance of DL models may vary due to algorithmic factors such as random seed, some researchers witness a consistent performance fluctuation among patients with diverse characteristics or what is referred to as \textit{sensitive attributes}.}
For instance, Stanley~\etal~\citep{stanley2022fairness} evaluated the disparities in {performance} and saliency maps in sex prediction using MR images, observing considerable discrepancies between \revise{White} and \revise{Black} children. Similarly, CheXclusion found that female, \revise{Black}, and low socioeconomic status patients were more likely to be under-diagnosed, as compared to their male, \revise{White}, and high socioeconomic status counterparts in chest X-ray datasets~\citep{seyyed2021underdiagnosis}.
This situation is not unique in medical image analysis (MedIA). Plenty of studies have addressed the existence of unfairness across multiple imaging modalities (MRI~\citep{puyol2021fairness,puyol2022fairness}, X-ray~\citep{seyyed2020chexclusion,seyyed2021underdiagnosis,glocker2021algorithmic}) and body parts (brain~\citep{ribeiro2022fair,ioannou2022study,petersen2022feature}, chest~\citep{cherepanova2021technical}, heart~\citep{puyol2021fairness,puyol2022fairness}, skin~\citep{xu2023fairadabn,mehta2023evaluating}), and different sensitive attributes (sex~\citep{larrazabal2020gender,petersen2022feature}, age~\citep{brown2022detecting,adeli2021representation}, race~\citep{zhang2018mitigating,deng2023fairness}, skin tone~\citep{li2021estimating,pakzad2022circle}).
Besides, this issue is also found in other healthcare applications where the inputs to the system are electronic medical records~\citep{yang2023algorithmic,yang2023adversarial} or RNA sequences~\citep{celeste2023ethnic}.

The phenomenon where the effectiveness of DL models notably favors {or opposes} one subgroup over another is termed \textbf{unfairness}~\citep{lara2023towards}. This issue is of profound ethical significance and necessitates urgent attention, as it contravenes fundamental bioethical principles~\citep{beauchamp2001principles}. Moreover, it poses substantial impediments {to} developing reliable and trustworthy DL systems for clinical applications~\citep{liu2023translational}.
{
One thing that must be taken in mind is that current studies about fairness mainly focus on the mathematical notions of fairness, i.e. using pre-defined metrics to measure the degree of unfairness.
However, this form of definition does have shortcomings, as it can only deal with correlations rather than causation, leading to challenges in understanding how fairness statistics are derived and attributing responsibility~\citep{srivastava2019mathematical}.
Besides, as the societal definitions of fairness evolve and are perceived differently by individuals, it is hard to find a proper formula to describe fairness consistently~\citep{jones2010building}.
}

{Failure to adequately address fairness issues could result in a subgroup of patients receiving inaccurate or under-diagnoses~\citep{green2003unequal,anderson2009racial,seyyed2021underdiagnosis}, potentially leading to deterioration and causing lifelong harm to the patients.} Clinicians may face difficulties in placing trust and confidently integrating deep learning methods into their routine practices. 
Recently, several studies in medical areas have urged the assessment of fairness in MedIA, including surgery~\citep{mittermaier2023bias}, nuclear medicine~\citep{currie2021ethical}, and dental care~\citep{batra2023new}, which shed {light on} this area. Nonetheless, owing to distinct research focuses and {various} scopes in comprehending issues among AI scientists and clinicians, it is imperative to establish a bridge for understanding fairness between these two groups~\citep{liu2023translational}.

To this end, we have undertaken a systematic review aimed at addressing fairness concerns within DL-based MedIA. This review endeavors to introduce fundamental fairness concepts while categorizing existing studies about fair MedIA. Our aspiration is that this comprehensive review will aid both AI scientists and clinicians in understanding the present landscape and necessities concerning fair MedIA, thereby fostering the advancement of fair medical AI.

\subsection{The Basics of Group Fairness}
\input{tablefc.tex}

Fairness, as a concept describing collective societal problems, has been discussed widely throughout the development of human society~\citep{liu2023translational}. Although the definition of fairness varies in different areas, the gist is the same, i.e. all citizens should have the right to be treated equally and equitably.
In AI research, fairness can be categorized into individual fairness~\citep{dwork2012fairness}, group fairness~\citep{barocas2017fairness}, max-min fairness~\citep{lahoti2020fairness}, counterfactual fairness~\citep{kusner2017counterfactual}, etc.
Among them, \textit{group fairness} is used by most of the studies in DL-based MedIA. Thus, in this section, we present the basics of group fairness for better comprehension.

Group fairness requires that the DL model should have equal utilities for all the subgroups in the test set.
Specifically, supposing a scenario where each subject in the dataset is a triplet, i.e. $S_i = \{X_i, Y_i, A_i\}$, where $X_i$ denotes the image data of subject $S_i$, $Y_i$ denotes the target label, and $A_i$ denotes the \textit{protected sensitive attributes}.

\begin{figure}
    \centering
    \includegraphics[width=1.0\linewidth]{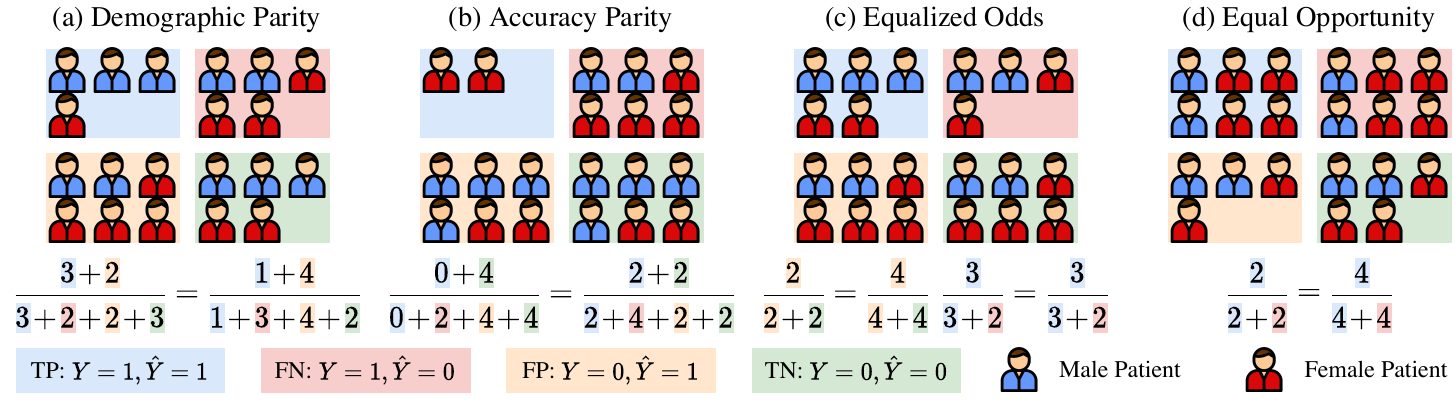}
    \caption{{Ideal situations where various fairness criteria are satisfied. From left to right: Demographic Parity, Accuracy Parity, Equalized Odds, Equal Opportunity. The equations below compute the value of different criteria for the Male and Female groups.}}\label{fig:ideal_fairness}
\end{figure}

The general pipeline of group fairness evaluation is as follows. {\it First}, the test set is split into mutually exclusive subgroups by the sensitive attribute $A_i$.
In MedIA, sensitive attributes can be information about the patients, including age, sex, race, skin tone, blood type, handedness, BMI, etc.
{\it Then}, several group-wise fairness metrics are computed over each subgroup. The commonly used fairness criteria are shown in Table~\ref{tab.fc}. For better understanding, we simulate four toy phenomenons where one of these fairness criteria is satisfied in Fig.~\ref{fig:ideal_fairness}. {For example, as shown in Fig.~\ref{fig:ideal_fairness}~(a), ideal demographic parity requires the Male and Female groups to have equal probability to be predicted as illness, i.e. $P(\hat{Y}=1\mid A=0) = P(\hat{Y}=1\mid A=1)$.} {However, these fairness indicators could contradict each other and might not be satisfied at the same time. For example, as shown in Fig.~\ref{fig:ideal_fairness} (a), the demographic parity is achieved. However, the accuracy on the Male group, i.e. $\text{ACC}_{\text{Male}} = P(\hat{Y}=Y\mid A=0) = \frac{3+3}{3+2+2+3}=\frac{3}{5}$, while the accuracy on the Female group, i.e. $\text{ACC}_{\text{Female}} = P(\hat{Y}=Y\mid A=1) = \frac{1+2}{1+3+4+2}=\frac{3}{10}$, which means that the Accuracy Parity is not satisfied.} Thus, it is important to select proper fairness criteria based on specific tasks~\citep{mbakwe2023fairness}.
{\it Finally}, the disparity between group-wise fairness metrics is computed to judge the overall fairness. The measurement function could be subtraction~\citep{bird2020fairlearn}, division~\citep{bird2020fairlearn}, SER~\citep{puyol2021fairness}, STD~\citep{wang2020mitigating}, NR~\citep{pakzad2022circle}, etc.

\begin{figure}
    \centering
    \includegraphics[width=1.0\linewidth]{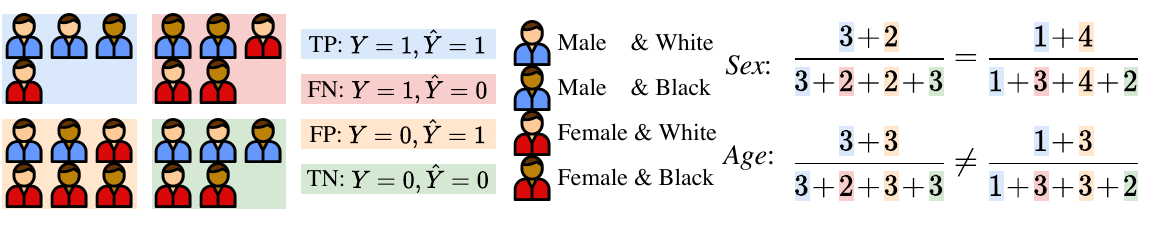}
    \caption{{In a scenario involving two sensitive attributes, namely \textit{sex} (male, female) and \textit{race} (White, Black), demographic parity is achieved concerning \textit{sex} but not \textit{race}. }}\label{fig:multisa}
\end{figure}

{Compared to individual fairness~\citep{dwork2012fairness}, which requires similar outputs for similar samples and is easy to be affected by the outlier samples, group fairness stabilizes the fairness analysis by group-wise average measurements.}
But also has shortcomings as the model might satisfy fairness constraints in one grouping scheme while being unfair in another group scheme, resulting in an ambiguous conclusion.
{For example, Fig.~\ref{fig:multisa} shows a situation where the patients can be split into subgroups based on \textit{sex} and \textit{race} (White / Black). Although DP is satisfied on \textit{sex}, i.e. $P(\hat{Y}=1 \mid A=\text{Male}) = \frac{5}{10} = P(\hat{Y}=1 \mid A=\text{Female}) $, DP is not satisfied on \textit{race}, i.e. $P(\hat{Y}=1 \mid A=\text{White}) = \frac{6}{10} \neq \frac{4}{9} = P(\hat{Y}=1 \mid A=\text{Black}) $}.

\section{Results}

\begin{figure}
    \centering
    \includegraphics[width=0.6\textwidth]{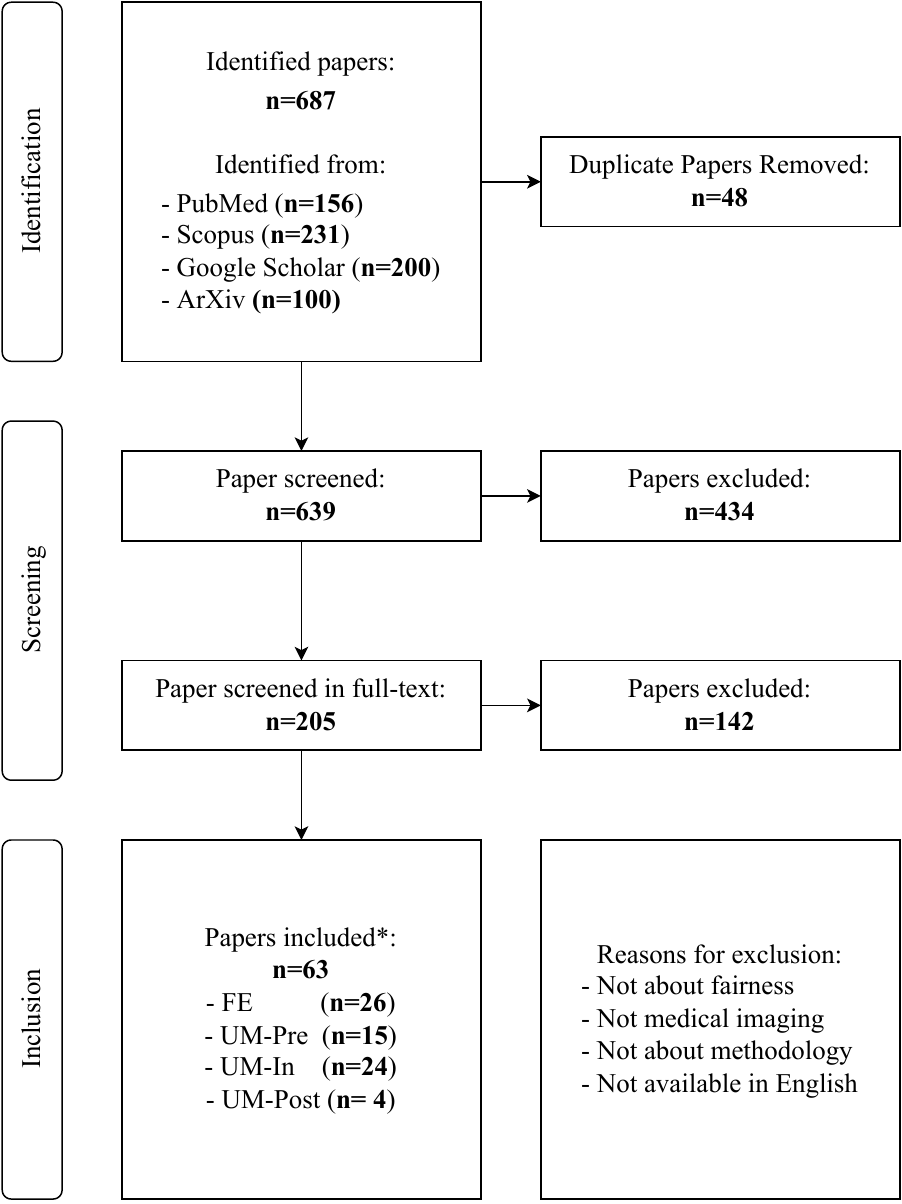}
    \caption{{PRISMA diagram for this review. * denotes that six studies have been overcounted due to their involvement in research across multiple directions FE: Fairness Evaluation; UM: Unfairness Mitigation; Pre: Pre-processing; In: In-processing; Post: Post-processing.}}
    \label{fig:prisma}
\end{figure}

\begin{figure}
    \centering
    \includegraphics[width=0.8\textwidth]{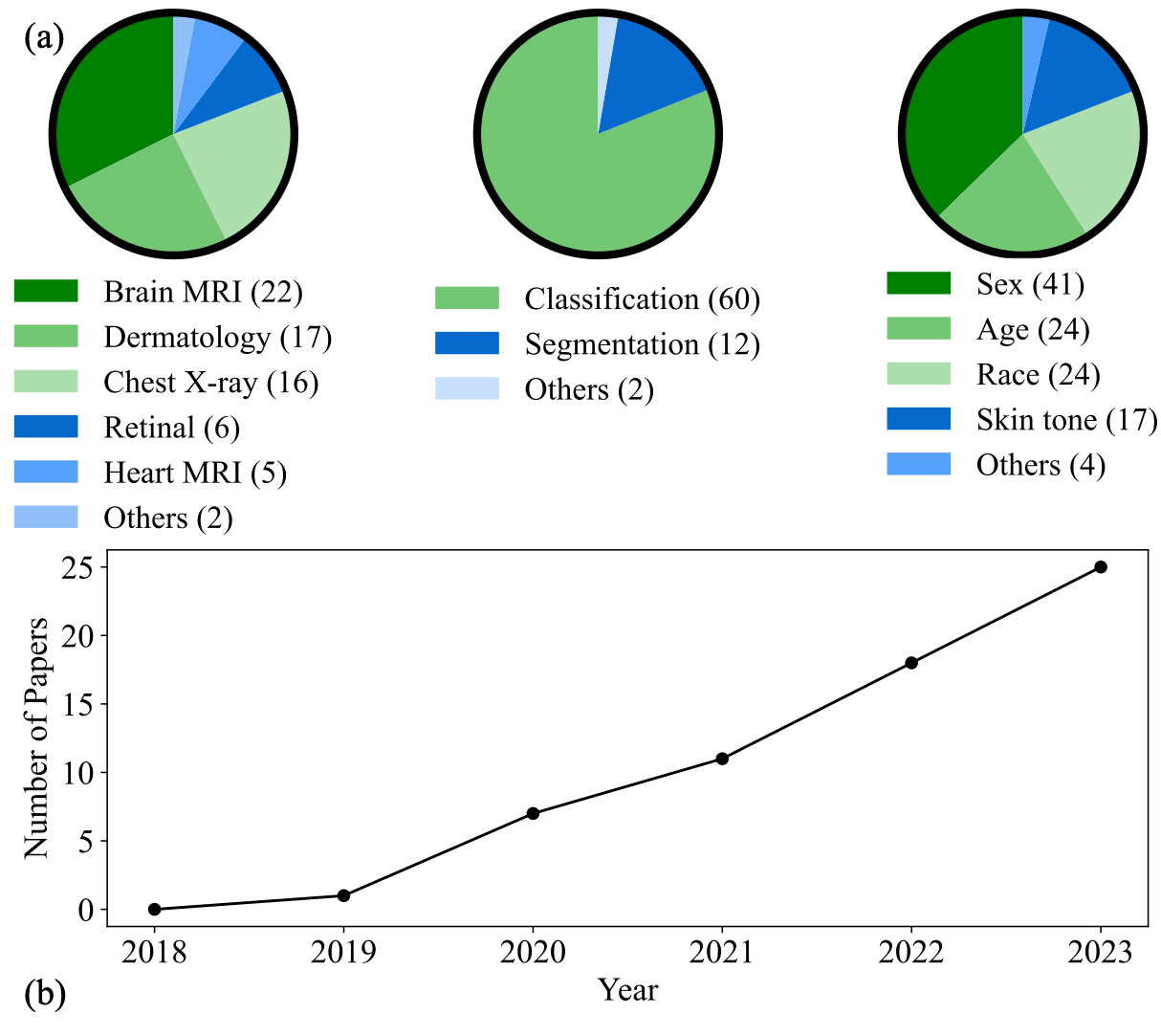}
    \caption{{Summary of data extracted from studies in our systematic review: (a) Annual trends in research on fairness in MedIA. (b) Prevalence of various medical imaging modalities, research tasks, and associated sensitive attributes.}}
    \label{fig:barplot}
\end{figure}

A total of 687 papers were identified by our systematic research. After removing duplicates and irrelevant studies based on our criteria (see Methods), 63 studies were included for information extraction and categorization.
Fig.~\ref{fig:prisma} presents the flowchart of this review based on PRISMA. Fig.~\ref{fig:barplot} and Table~\ref{tab:extract_data} describe the statistics of the extracted data.

\input{tableed.tex}

{
From Fig.~\ref{fig:prisma} and Fig.~\ref{fig:barplot} we can find that research about fairness in MedIA began in 2019 and the annual number of publications grew about 6 to 7 per year.
Fairness in MedIA is mainly assessed on Brain MRI, Dermatology, and Chest X-ray. This is easy to understand as the development of the brain is highly related to sex and age, while the skin part usually appears in dermatology images, which may lead to spurious relations for diagnosis. Chest X-ray, however, has the largest amount of samples compared to other modalities, which provides a potential for evaluating fairness.
Besides, most of the current research \revise{focuses} on classification and segmentation. Some attempts have also been conducted on anomaly detection~\citep{bercea_bias_2023} and regression~\citep{mehta2023evaluating}.
As for the sensitive attributes, sex, age, race, and skin tone are the most concerned.
}

Specifically, the research about fairness in MedIA mainly consists of two folds. One aims at benchmarking fairness in various medical tasks and discovering the mechanism behind unfair performances, and the other category attempts to mitigate unfairness in current applications.
{The studies about fairness evaluation pay more attention to the existence of unfairness in various medical applications. While unfairness mitigation via in-processing strategies is more often considered (24 / 43).}
Note that some studies include research in multiple directions thus we repeat counts in Fig.~\ref{fig:prisma}.

\subsection{Fairness Evaluation}

The starting point of addressing fairness issues in MedIA is to evaluate the existence of unfairness in MedIA tasks, by applying performance comparison among different sensitive groups.
On the other hand, some researchers try to discover visual patterns that account for unfair model performance and study the relationship between fairness and other concepts.

\subsubsection{Benchmarking Unfairness in Various MedIA Tasks}
Up to now, plenty of studies have been conducted on benchmarking unfairness in MedIA tasks. These studies evaluate diagnosis {performance} either on multiple network architectures or trained with different attribute ratios. 

These studies vary in body parts and modalities, including brain MR~\citep{stanley2022fairness,ribeiro2022fair,yuan_algorithmic_2023,wesarg_auditing_2023, klingenberg_higher_2023}, breast MR~\citep{huti_investigation_2023}, Mammography~\citep{schwartz2021association}, Chest {X-ray}~\citep{seyyed2021underdiagnosis,seyyed2020chexclusion,glocker2021algorithmic}, cardiac MR~\citep{puyol2022fairness}, head and neck PET/CT~\citep{salahuddin_head_2023}, and skin lesion~\citep{kinyanjui2020fairness,kalb_2023_revisiting}.
Not surprisingly, most of the studies notice significant subgroup disparities in the utility except for the study in~\citep{kinyanjui2020fairness}, which does not find an observable trend between model utility and skin tone \citep{kalb_2023_revisiting}.
Recently, researchers {expended} the range of tasks and the type of algorithms, such as regression~\citep{picarra_analysing_2023}, anomaly detection~\citep{bercea_bias_2023}, reconstruction~\citep{du_unveiling_2023}, multi-instance learning~\citep{sadafi_study_2023}, random forest~\citep{huti_investigation_2023}, and transformer~\citep{lee_investigation_2023}, which fills the gap in fair MedIA.
Besides, Zong~\etal~\citep{zong2022medfair} benchmark ten unfairness mitigation methods on nine medical datasets. However, they find that none of the ten methods outperforms the baseline with statistical significance.
Similar results are also found in CXR-Fairness~\citep{pmlr-v174-zhang22a}.

To go further, some research examines whether the subgroup ratio in the train set affects the performance of DL models. 
Larrazaba~\etal~\citep{larrazabal2020gender} train DL models on Chest X-ray datasets and find that the diagnosis {performance} between the female and the male is significantly different.
This phenomenon could result from the fact that the diagnosis for one subgroup is more difficult than that for another subgroup, due to different image quality~\citep{ganz2021assessing}.
Opposite to this finding, Petersen~\etal~\citep{petersen2022feature} evaluate the performance of DL models on the ADNI dataset with different subgroup ratios and notice that the performance {of} the under-represented group is not significantly different from that of the over-represented group.
Besides, Ioannou~\etal~\citep{ioannou2022study} examine fairness by comparing the average {Dice Similarity Coefficient (DSC)} on each spatial location. Interestingly, they find that although unfairness exists in some regions severely, in other regions the {DSC} shows little or no disparity, which prompts us that unfairness evaluation should be conducted on different components respectively rather than on the overall utility.

\subsubsection{Unfairness Source Tracing and Mechanism Discovery}

\begin{figure}
    \centering
    \includegraphics[width=1.0\textwidth]{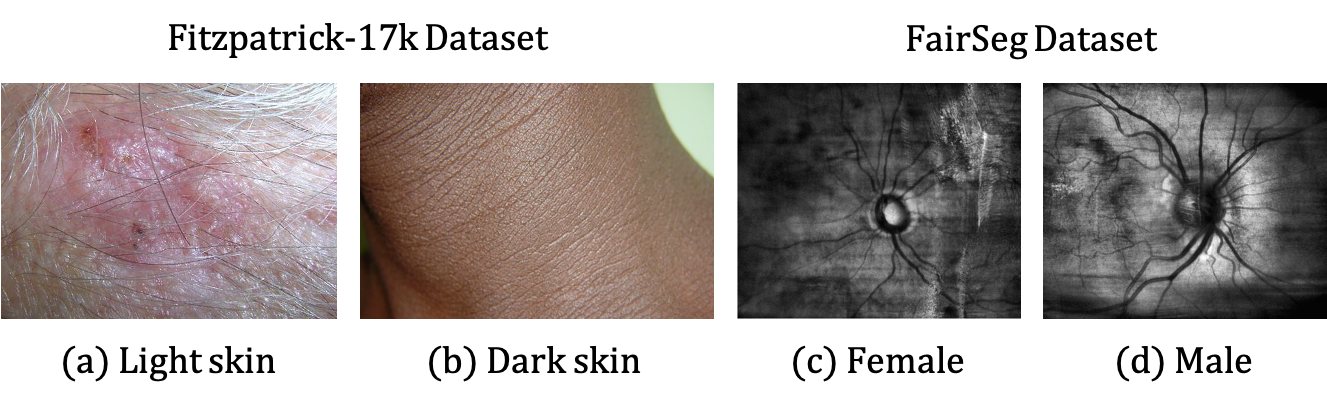}
    \caption{\revise{Visual disparities between images with different sensitive attributes. (a)(b): images with dark skin and light skin from Fitzpatrick-17 Dataset~\citep{groh2021evaluating}; (c)(d): images of a male patient and a male patient from FairSeg Dataset~\citep{tian2024fairseg}.}}
    \label{fig:visual}
\end{figure}
{To discover the source of unfairness, many studies focus on the visual disparities among samples with different attributes, which are recognized widely across different modalities~\citep{campos2014gender}, and suppose that unfairness comes from visual differences. For example, as shown in Fig.~\ref{fig:visual}, there is a huge visual disparity between the White and the Black, the male and the female, and the young and the old.} {Liang~\etal try to generate a sex-inverted X-ray image using a generative adversarial network and focus on the regions that have the largest disparity between the original image and its sex-inverted counterpart. By comparing the visual differences between the two images, they come up with some insights for understanding the source of unfairness and improving the DL models' interpretability.}
Similarly, Jimenez-Sanchez~\etal~\citep{jimenez-sanchez_detecting_2023} witness that the drain in the chest X-ray images may lead to shortcut learning for DL models, which causes unfairness.
On the other hand, Jones~\etal~\citep{jones2023role} notice that the ability of classifiers to separate individuals into subgroups is highly relevant to subgroup disparity.
Besides, Kalb~\etal~\citep{kalb_2023_revisiting} find that the level of unfairness varies when computing the {Individual Typology Angle (ITA)} of the image using different methods, which proves that unfairness comes from the label annotation procedure.
In chest X-ray diagnosis, Weng~\etal~\citep{weng_are_2023} suppose that the differences in the imaging quality of the breasts between the male and the female are the reason for unfairness. Although the result does not support this hypothesis, their research gives insights for inspecting the source of unfairness.
Moreover, Du~\etal~\citep{du_unveiling_2023} evaluate model fairness {in} MRI reconstruction tasks and find that the estimated Total Intracranial Volume and normalized Whole Brain Volume might be the cause of unfairness.

Uncertainty also has a relationship with fairness.
Lu~\etal~\citep{lu2021evaluating} try to compare several uncertainty measurements by evaluating subgroup disparity and find that although the aggregate utilities are similar, subgroup disparity varies among different uncertainty measurements.
However, the result of~\citep{mehta2023evaluating} conducted on three clinical tasks shows that although some unfairness mitigation methods have positive effects on fairness, they might harm the uncertainty of the model prediction.

\subsection{Unfairness Mitigation}
\begin{figure}
    \centering
    \includegraphics[width=1.0\textwidth]{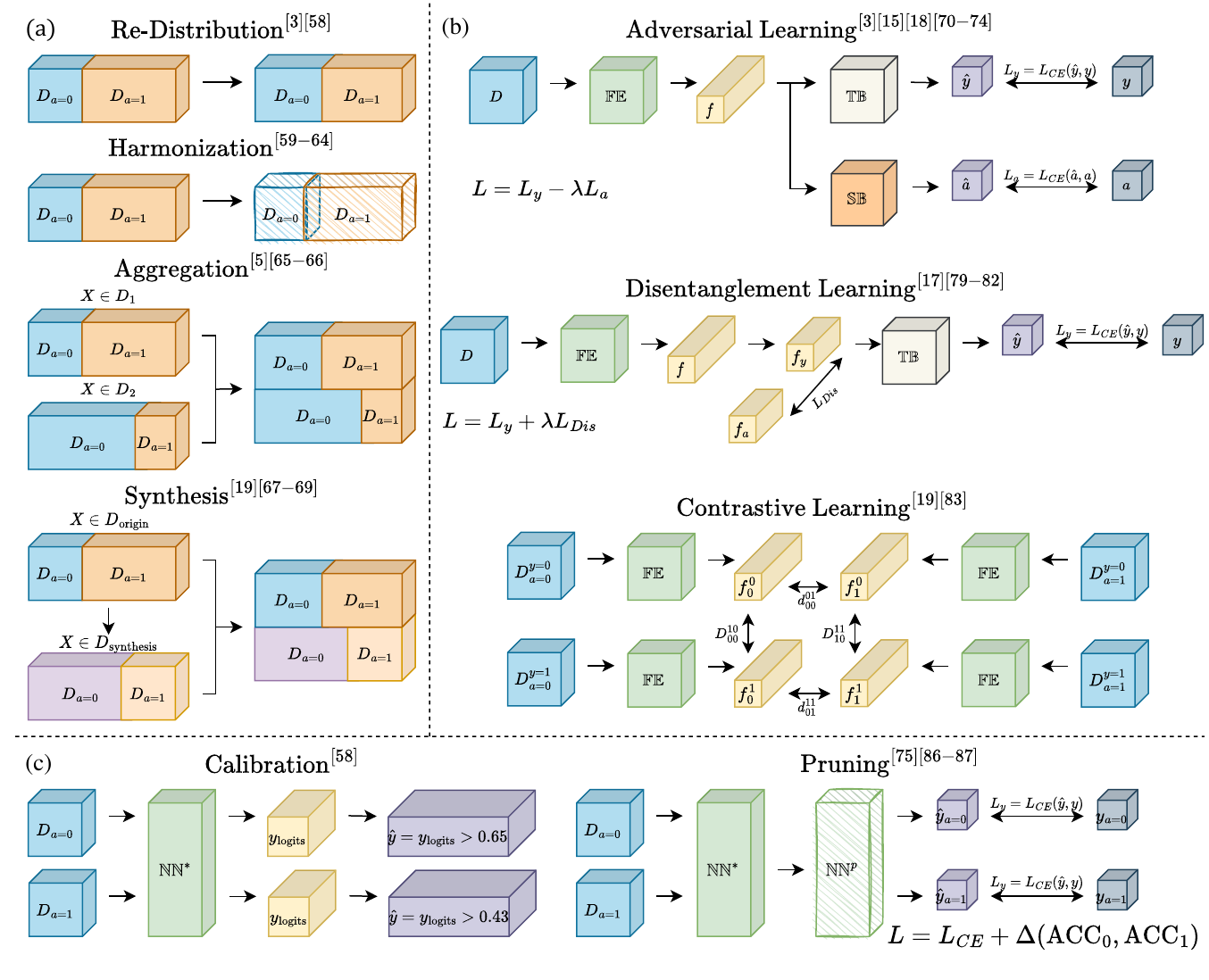}
    \caption{{Schematic diagram of unfairness mitigation algorithms. (a) Pre-processing methods. $D_1, D_2$: Two independent datasets; $D_{\text{origin}}, D_{\text{synthesis}}$: The original dataset and synthesized dataset. (b) In-processing methods. $\mathbb{FE}, \mathbb{TB}, \mathbb{SB}$: Feature Extractor, Target Branch, and Sensitive Branch, which are three parts of an adversarial network; $f$: latent feature vector; $y, a,\hat{y},\hat{a}$: the ground truth target task label, sensitive attributes, and their corresponding predictions generated by the neural network; $\mathcal{L}_{CE}$: Cross-Entropy loss, measuring the difference between the predicted label and the ground truth label; $\mathcal{L}_{\text{Dis}}$: Disentanglement loss, for example, MMD-Loss~\citep{borgwardt2006integrating}, measuring the distance between two distribution; $D_{00}^{10}$: requiring the maximum distance between $f_0^0$ and $f_0^1$; $d_{00}^{01}$: requiring the minimum distance between $f_0^0$ and $f_1^0$. (c) Post-processing methods. $\mathbb{NN}^*$: a pre-trained and fixed Neural Network; $y_{\text{logits}}$: predicted probability of $y$, range from 0 to 1; $\mathbb{NN}^p$: pruned $\mathbb{NN^*}$; $\Delta(\text{ACC}_0, \text{ACC}_1)$: difference between accuracy on subgroup test set $D_{a=0}$ and $D_{a=1}$.}}
    \label{fig:schematic diagram}
\end{figure}

According to~\citep{xu2023fairadabn}, strategies aiming at unfairness mitigation can be categorized into pre-processing, in-processing, and post-processing. The schematic diagram of each category is shown in Fig.~\ref{fig:schematic diagram}.

\subsubsection{Pre-Processing}
The pre-processing methods mainly \revise{focus} on data modification or remedy, which can be categorized into data re-distribution, harmonization, aggregation, and synthesis.

\textbf{Re-distribution} addresses unfairness by adjusting the balance of subgroups.
This can be achieved by either resampling the train set or by controlling the number of samples in different subgroups within each mini-batch. By ensuring a more balanced representation of subgroups, re-distribution can help to mitigate unfairness in DL models.
Puyol-Ant\'{o}n~\etal~\citep{puyol2021fairness} adopt stratified batch resampling on the baseline model and notice a significant improvement in fairness. This strategy is also proved to be efficient in~\citep{oguguo2023comparative}.

Data \textbf{harmonization} removes sensitive information from the input images. 
This can be conducted by using segments or bounding boxes to remove the skin part from dermatological images~\citep{bissoto2019constructing,bissoto2020debiasing}, enhancing lesion boundary using image processing method~\citep{yuan_edgemixup_2022}, applying Z-score normalization or ComBat algorithm~\citep{wachinger2021detect}, or using differential privacy methods~\citep{wesarg_-identification_2023}.
Recently, Yao~\etal~\citep{yao2022improving} use a language-guided sketching model to transform the input images into sketches. Their result shows less disparity among subgroups, which indicates the potential of data harmonization in mitigating unfairness.

Data \textbf{aggregation} mitigates unfairness using information introduced by external datasets. 
The external dataset could be in the same modality~\citep{seyyed2020chexclusion,wang_bias_2023} or different modalities~\citep{zhou2021radfusion}. 
For example, Zhou~\etal~\citep{zhou2021radfusion} adopt the information of electronic health record data via an ElasticNet and improve the fairness of the model on pulmonary embolism detection by multi-modal fusion.

Data \textbf{synthesis} uses generative models to synthesize new samples to increase the number of training samples and balance the subgroup ratio in the training set.
This can be done by either generating new samples with the same sensitive attribute as the original samples while varying in target label~\citep{joshi2021ai} or by generating new samples with opposite sensitive attributes but preserving {the} target label~\citep{burlina2021addressing,pakzad2022circle,pombo_equitable_2023}.

\subsubsection{In-Processing}
The in-processing methods focus on adjusting the architecture of models or adding extra losses or constraints to reduce biases among subgroups.

The \textbf{adversarial architecture} is the most common in-processing method, which adds an adversarial branch to the original architecture to minimize the influence of sensitive information in the latent space by using a gradient reversal layer as the adversarial branch~\citep{zhao2020training,adeli2021representation,abbasi2020risk,bevan2022detecting,barron2020generalization,puyol2021fairness,yang2023adversarial,correa_systematic_2022,stanley2022fairness}. The biggest difference between them is the choice of loss functions for sensitive attribute prediction, 
One step further, Li~\etal~\citep{li2021estimating} add an extra branch except for the adversarial one, which predicts the degree of fairness on the test set without knowing the sensitive attributes.

Another category of in-processing methods directly adds \textbf{fairness-related constraints} to the optimization objective.
The constraints {include} GroupDRO~\citep{oguguo2023comparative}, differentiable proxy functions of fairness metrics~\citep{marcinkevics2022debiasing,cherepanova2021technical}, bias-balanced Softmax~\citep{luo2022pseudo}, and margin ranking loss~\citep{lin_improving_2023}.
However, as shown in~\citep{cherepanova2021technical}, this type of method can lead to over-parametrization and overfitting, resulting in a fluid decision boundary that can lead to fairness gerrymandering~\citep{kearns2018preventing}.
{i.e. the DL models might be fair on \textit{sex} and \textit{age}, respectively, but are unfair when evaluated on the \textbf{combination} of \textit{sex} and \textit{age}.}

\textbf{Disentanglement learning} extracts the feature vector from the input image and projects the feature vector into a task-agnostic portion and a task-related portion. 
This can be achieved by either maximizing the entropy of the task-agnostic portion~\citep{sarhan2020fairness} or by maximizing the orthogonality between the task-agnostic and the task-related portions~\citep{deng2023fairness}.
Some studies regard the relationship between the two portions as a linear model.
For example, Vento~\etal~\citep{vento2022penalty} introduce the metadata into the model and describe the relationship using a general linear function, i.e. $\textbf{f}=\textbf{M}\beta + \textbf{r}$, where $\textbf{f}$ is the origin feature vector, $\textbf{M}$ is the metadata, $\beta$ is the learnable coefficient, and $\textbf{r}$ is the residual task-related portion. 
A similar method is also used in~\citep{more2021confound} to mitigate the confounding bias in an fMRI dataset, where the coefficient $\beta$ is estimated either on the whole dataset or each fold of the training set.
The relationship could also be simply joint, as shown by Aguila~\etal~\citep{lawry2022conditional}, who train a conditional variational auto-encoder on structural MRI data to disentangle the effect of covariates from the latent feature vectors.

The intuition behind \textbf{contrastive learning} is that in the latent feature space, the distances among feature vectors belonging to the same target class and different sensitive attributes should be minimized, whereas the distances among feature vectors from different target classes and the same sensitive attribute should be maximized.
For example, Pakzad~\etal~\citep{pakzad2022circle} use a skin tone transformer based on StarGAN to transform the skin tone of the input image and then regularize the $L_2$ norm between the feature vectors extracted from the shared feature extractor.
Furthermore, Du~\etal~\citep{du2022fairdisco} implement a contrastive learning schema by adding an extra contrastive branch constraining the projected low-dimension feature vector.

There are some other attempts to mitigate unfairness in the in-processing procedure.
Inheriting the idea of domain adaptation, FairAdaBN~\citep{xu2023fairadabn} 
reduces unfairness by adding extra adaptors to the original model which can adaptively adjust the mean and variance of the feature vector according to the sensitive attribute. 
Fan~\etal~\citep{fan2021fairness} design a special federated learning setting, where each client in the swarm learning only consists of samples with the same sensitive attribute. 
{Moreover}, FairTune~\citep{dutt2023fairtune} tries to mitigate unfairness by using parameter-efficient fine-tuning from large-scale pre-trained models. By considering fairness, it can improve fairness in a range of medical image datasets.

\subsubsection{Post-Processing}

The post-processing methods mitigate unfairness by processing the fixed models by calibrating the output of DL models or pruning the model's parameters.

\textbf{Calibration} uses different prediction thresholds for each subgroup to achieve fairness. Oguguo~\etal~\citep{oguguo2023comparative} apply the reject option classification algorithm, which believes that unfairness occurs around the decision boundaries. It sets an outcome-changing interval where the output of the unprivileged group is re-modified. The calibration method is one of the most {useful} mitigation methods as it can be easily adopted on multiple DL models.

\textbf{Pruning} tries to satisfy fairness criteria by distilling the model's parameters.
In their recent work, Wu~\etal~\citep{wu2022fairprune} compute the saliency of each neuron in the network and utilize a pruning strategy to remove features associated with a specific group. This approach helps prevent sensitive information from being encoded into the network. {Moreover}, Marcinkevics~\etal~\citep{marcinkevics2022debiasing} also use a pruning strategy to deal with unfairness, where the sensitive attribute is unknown. To go further, Huang~\etal~\citep{wesarg_mitigating_2023} extend the neuron importance measurement proposed in~\citep{marcinkevics2022debiasing}, achieving better fairness while reserving model performance.

\subsection{Fairness Datasets in MedIA}

To prompt fairness research in MedIA, we collect publicly available datasets with sensitive attributes in Table~\ref{dataset_with_attrs}, which are categorized by image modality type, task type, attributes type, and the number of images in each dataset. We hope this Table can benefit the late-comer to find proper data to evaluate their algorithms.

\input{tabledwa.tex}
\input{tableglossary.tex}

\section{Discussion}

This review reveals the importance of assessing fairness in deep learning-based medical image analysis. 
We describe the basics of group fairness and categorize current studies in fair MedIA into fairness evaluation and unfairness mitigation.
In a word, fairness evaluation and unfairness mitigation is a rapidly growing and promising research field in MedIA, for AI scientists, clinicians, and the power-holder. We must pay more attention to this area to build a fair society for citizens of different sexes, ages, races, and skin tones.
Here, we discuss the challenges and opportunities for improving fairness in MedIA.

\subsection{Being aware of the Sources of Unfairness to Find Corresponding Solutions}

When adopting AI algorithms in medical applications, clinicians must be aware that, during the development of AI algorithms, most of the methods only focus on the diagnosis performance while ignoring whether the algorithm has biased or unfair utilities towards different subgroups.
As a result, clinicians may find that the algorithm tends to underdiagnose specific sub-populations, and the corresponding treatment for each group might also differ. This phenomenon will make the clinician confused about clinical decision-making and wonder why the AI algorithms perform unfairly like that.

Thus, it is important to discover the sources of unfairness in MedIA, not only to help clinicians to understand potential biases in the AI algorithms and propose reliable treatment but also to help the AI scientists be aware of the source of unfairness and produce corresponding solutions and design better AI products. 
Moreover, after clinicians and AI scientists are aware of the sources of unfairness in each application, the government can devise actions focusing on the root reason and prevent the recurrence of unfairness in the future.

\begin{figure}
    \centering
    \includegraphics[width=1.0\textwidth]{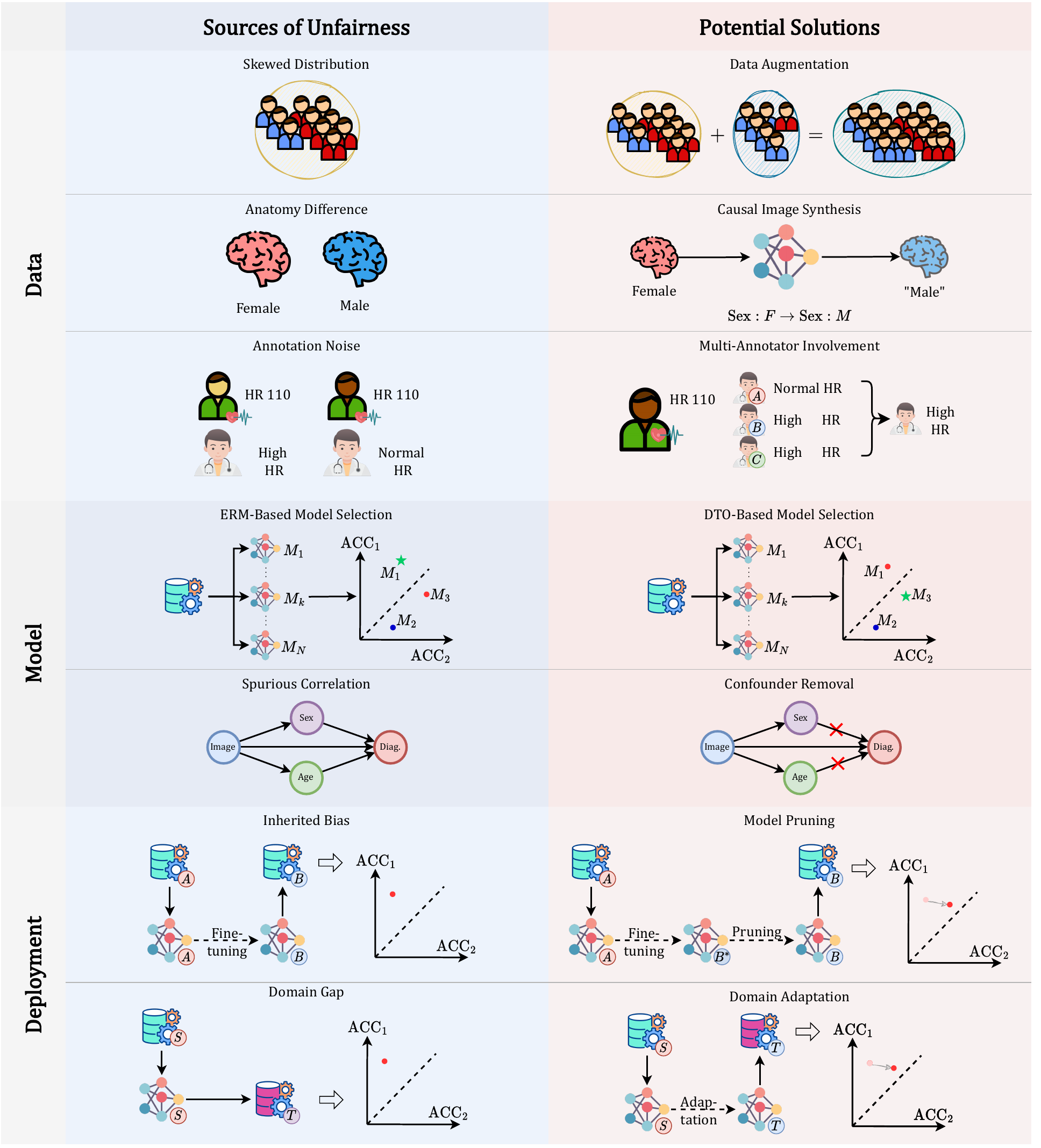}
    \caption{{Sources of unfairness and potential solutions. From top to bottom: skewed data distribution $\rightarrow$ aggregate data from multiple datasets; anatomy difference between subgroups $\rightarrow$; annotation differences for each subgroup $\rightarrow$ using causal image synthesis methods to transfer the input to the synthesis ones with opposite attribute; annotation noise $\rightarrow$involve multi-annotators to stabilize annotation; ERM-based model selection which chooses models with the highest overall performance $\rightarrow$ DTO-based model selection which consider both performance and fairness; spurious correlations between sensitive attributes and diagnosis $\rightarrow$ removing the effects of the confounder; inherited bias from the pre-train dataset $\rightarrow$ pruning the pre-trained with fairness constraints; domain gaps between the source dataset and target dataset $\rightarrow$ using domain adaptation methods to transfer models.}}
    \label{fig:sources}
\end{figure}

Following~\citep{ricci_lara_addressing_2022}, we categorize the sources of unfairness based on the components of a DL system pipeline, i.e. data, model, and deployment. Fig.~\ref{fig:sources} illustrates the schema of these sources and potential solutions.

Unfairness existing in the \textbf{data} consists of three parts. 
\textit{First}, the skewed data distribution of the diagnosis label and the sensitive attributes might cause the insufficient feature representation of the specific group~\citep{clark2013cancer,puyol2022fairness}. 
\textit{Second}, due to the anatomical differences between different subgroups, the difficulty for DL models to provide precise diagnosis may also vary~\citep{webb2022addressing}. 
\textit{Third}, as the clinicians may have different annotation preferences on patients of different attributes~\citep{gonzalez2021disparities}, this annotation inconsistency will confuse the DL model and cause unfairness.

The DL \textbf{model} itself also causes unfairness. 
The general training process of DL models aims to find the model with the highest overall performance. However, a higher performance might lead to larger subgroup gaps~\citep{zong2022medfair}.
Besides, as DL models tend to learn easier but irrelevant correlations between the input image and the output diagnosis, they may attempt to produce the diagnosis based on spurious correlations rather than true medical evidence~\citep{zong2022medfair}.
Moreover, some DL models also amplify the unfairness that exists in the data and enlarge the fairness gaps among subgroups~\citep{wang2021directional}.

With the wide use of pre-trained DL models, unfairness also comes from the \textbf{deployment}.
Some pre-trained models inherit bias from the pre-training datasets and perform unfairly on the downstream tasks.
On the other hand, the non-neglectable domain gaps among the situations where the model is developed and deployed also affect the performance of the model and cause unfairness.

\subsection{Differences between Fair MedIA and Fair Facial Recognition}

The Facial recognition (FR) community was aware of unfairness issues, even earlier than the MedIA community~\citep{karkkainen2021fairface}. Research in fair FR aims to achieve equal TPR and FPR of recognition among subgroups with different sex, age, hairstyle, emotion, etc. 
Over the past years, several attempts have been made to align fair MedIA with fair FR, by transferring unfairness mitigation algorithms from FR to MedIA~\citep{zong2022medfair,pmlr-v174-zhang22a}. However, experiments prove that most of the useful methods in fair FR are not applicable in fair MedIA, which attracts people's attention to the differences between the two areas.
Table~\ref{tab:FR_align} illustrates the comparison between fair FR and fair MedIA.

\input{tablefr.tex}

The largest difference between fair FR and fair MedIA is the variation of image modalities.
As most of the images in FR are RGB images, the input in MedIA varies from 2D X-ray images, dermoscopy, and mammography to 3D MRI, and PET/CT. The huge disparity among the multi-modality input and complex types of tasks (classification, segmentation, detection) makes it impossible to find a common solution for all tasks.
Besides, the amount of samples in the MedIA dataset is several orders of magnitude smaller than those in facial datasets, which brings more difficulties for robust feature representation.
The type of sensitive attribute also varies. While attributes in fair MedIA mainly describe the demographics of a patient, attributes in fair FR focus more on the \revise{appearance} of a person. This disparity affects the difficulty of attribute recognition.
As a result, the composition of MedIA datasets is usually skewed, due to the different morbidity across subgroups. In contrast, there are several specially designed face datasets with balanced attributes for fair FR, such as FairFace~\citep{karkkainen2021fairface}.
The lack of balanced medical datasets seriously hinders the benchmark of unfairness mitigation algorithms in fair MedIA.

\subsection{Mitigating Discrepancies in the Interpretation of Fairness across Diverse Perspectives}
While AI scientists and clinicians have their understanding of fairness, the gaps between the mathematical definitions and the dilemma in medical scenarios cannot be neglected to achieve health equity.

Most of the current research in AI fairness is conducted on measuring and mitigating unfairness with the mathematical form of fairness, i.e. the numeric differences between manually designed metrics such as DP, AP, EqOdd, etc.
However, from the scope of clinicians, the \textit{equality} in numbers does not always mean the \textit{equity} of treatment~\citep{mccradden2020ethical}. In other words, the performance disparity does not equal unfairness. Forced insistence on numerical equality will destroy the causal relationship between the patients' metadata (sensitive attributes) and the diagnosis outcome.

\revise{To address fairness in AI, we need to decide which sensitive attributes should be evaluated. Generally, AI scientists prefer to analyze the statistical relationship between the target task and the sensitive attribute, i.e. does the model have a biased outcome on this attribute? On the contrary, clinicians pay more attention to the physiological causality between the two terms, for example, will the anatomical difference between the male and the female affect the diagnosis difficulty? This different paradigm of attribute selection also brings gaps between AI scientists and clinicians when they address fairness in MedIA together~\citep{ricci_lara_addressing_2022}.}

The choices of fairness metrics also vary between AI fairness and clinical fairness.
While AI scientists adopt metrics that are derived from the confusion metrics, clinicians regard some of the metrics as unreasonable.
For example, one of the most commonly used metrics in AI fairness, demographic parity, requires that the patients from each subgroup should have the same probability of being predicted as ailing, i.e. $P(\hat{Y}=1 \mid A=0) = P(\hat{Y}=1 \mid A=1)$. However, in practical medical scenarios, where many illnesses are proven to be related to age or sex, the requirement of the same subgroup morbidity is not matched with reality.
Furthermore, the diagnosis difficulty may vary across subgroups due to anatomical differences~\citep{fairbairn2020sex}. In this situation, it is more appropriate for the DL algorithm to have an unequal diagnosis precision due to reasonable medical prior.

Moreover, due to the randomness of DL models, it is nearly impossible to have exactly the same performance among different subgroups. However, in medical applications, a slight fluctuation \revise{in} the subgroup diagnosis performance is acceptable due to the complexity of the illness. Thus, we need to carefully decide that \textit{what level of numerical difference means unfairness.} 
Furthermore, unlike utility metrics such as accuracy and area under the receiver-operating curve (AUROC), fairness metrics usually \revise{fluctuate} along the training procedure and are hard to converge. Therefore, additional efforts need to be made to assess the level of fairness properly.

\subsection{When Fairness Meets Large-scale Foundation Models}

\color{black}
Recently, foundation models, such as Large Language Model (LLM)~\citep{lee2020biobert,MedicalGPT}, Contrastive Language Image Pre-training (CLIP) and its variants~\citep{radford2021learning,eslami2023pubmedclip,lin2023pmc,lai2024carzero}, and Segment Anything Model (SAM) and its variants~\citep{kirillov2023segment,MedSAM,quan2024slide,wang2023sammed3d} have attracted people's attention by their superior zero-shot or few-shot performances on downstream tasks in medical applications.
However, as the training sets of these foundation models are usually inaccessible, it is important to ensure whether these models have unbiased utilities on different subgroups before adopting them in healthcare applications.

For example, LLMs, a category of DL models trained with countless corpus from all over the world, might introduce unfairness from the pre-training tasks, or perform unfairly due to the large domain gap between the pretraining task and the fine-tuned downstream tasks (unfairness from deployment)~\citep{li2023survey}.
Similarly, unfair performances are also witnessed in CLIP models, which try to align semantic information between text and image data to construct an excellent feature extractor~\citep{Wang2021MitigateGenderBiasInImageSearch,berg2022prompt}. This unfairness might be due to the spurious relations between sensitive attributes and the target label~\citep{hendricks2018women}.
SAM is another family of foundation models that focuses on image segmentation tasks, mainly with the help of point or box prompts.
As shown in the original SAM paper, the training images of SAM have a strong region preference, i.e., the number of images collected in Europe is larger than that in South America.
This bias in geographic distribution is propagated in the fine-tuned version of SAM in medical applications. 

In short, most of the large-scale foundation models and language models suffer from different levels of unfairness, due to domain gap, annotation noise, spurious correlation, or inherited bias from the training set. This phenomenon hurts the trustworthiness of DL-based algorithms and must be handled properly.
However, due to the huge amount of parameters of these foundation models, it is hard to mitigate unfairness using the aforementioned techniques, as the retraining of foundation models requires heavy computation and a long time.
Thus, it is crucial to come up with unfairness mitigation methods aimed at foundation models, including perturbing the feature space~\citep{xu2024apple} and editing the input images~\citep{jin2024universal}. 
\color{black}


\subsection{Cooperation is Required to Ensure Fairness in Medical Applications}

Addressing fairness in MedIA requires in-depth cooperation among AI scientists, ethicists, and clinicians.
In this cooperation, AI scientists should be aware of the limitations of the mathematical form of fairness with the help of ethicists and clinicians, and try to develop new algorithms that can mitigate unfairness effectively.
The clinicians could discover the causal relations between illness diagnosis and the metadata of the patients, finding out which type of disparity should be regarded as differences rather than unfairness.
Besides, the government can also incorporate fairness considerations into clinical AI guidelines to improve the awareness of fairness in the whole pipeline~\citep{collins2021protocol}.
By improving the comprehension of the operation mechanism of the DL models and the clinical context grasping, the researchers can update existing AI governance schemes to promote fairness.
While establishing multi-directional communication across various professions may present challenges, it is imperative to exert considerable effort toward addressing fairness in MedIA and safeguarding health equity on a global scale.

\section{Methods}

This review was conducted based on the PRISMA guidelines~\citep{takkouche2011prisma}.

\subsection{Search Strategy}
All the candidate articles were collected by a comprehensive search of 4 databases, including Scopus, PubMed, ArXiv, and Google Scholar using the query conditions: 
$\text{`medical fairness'} \mid \text{`fairness in medical image analysis'} \mid \text{`fair medical machine learning'} \mid `\text{`fairness in healthcare'}$. We limit the publish time from 2015 to 2023 for Scopus and PubMed, and only include the top 200 / 100 items for ArXiv and Google Scholar, respectively.

After removing duplicated papers, we included all research papers in English.
Since this review mainly focuses on methodology, we included papers describing methods for fairness issues in medical image analysis using deep learning algorithms.

For better categorization, we hierarchically split the papers into two classes, fairness evaluation and unfairness mitigation. Each of them was separated into more precise sub-areas accordingly.
Besides, we also reported the image modality and datasets used in their paper, tasks and sensitive attributes they worked on, and metrics about fairness they adopted.

\subsection{Data Extraction}
For the included studies, we extract information from the following aspects: (1) year of publication; (2) image modality; (3) datasets used; (4) type of task; (5) research area according to taxonomy; (6) sensitive attributes assessed on; (7) metrics used for fairness evaluation. By extracting these data, we hope to help the researcher better understand the routine of new studies about fair MedIA and come up with insights into both fairness evaluation and unfairness mitigation for MedIA.


\section*{Data Avaliability}
The authors declare that all data supporting the findings of this study are available within the paper and its Supplementary Information files.

\section*{Code Avaliability}
The authors declare that all code supporting the findings of this study are available within the paper and its Supplementary Information files.

\section*{Acknowledgements}

We thank Yongshuo Zong from the University of Edinburgh for his constructive suggestions.
This paper is supported by the Natural Science Foundation of China under Grant 62271465, Suzhou Basic Research Program under Grant SYG202338, and Open Fund Project of Guangdong Academy of Medical Sciences, China (No. YKY-KF202206).

\section*{Author Contributions}
Z.X, J.L., and S.K.Z designed the study.
Z.X. developed the search strings and analyzed the data.
Z.X and J.L. wrote the first draft of the manuscript.
All authors revised the manuscript and approved the final version of the submitted manuscript. 
The corresponding author attests that all listed authors meet authorship criteria and no others meeting the criteria have been omitted.

\section*{Competing Interests}
The authors declare no competing interests.

\bibliographystyle{IEEEtran}
\bibliography{AbbrBib}

\section*{Figure Legend}

\noindent\textbf{Fig.~\ref{fig:ideal_fairness}:} Ideal situations where various fairness criteria are satisfied. From left to right: Demographic Parity, Accuracy Parity, Equalized Odds, Equal Opportunity. The equations below compute the value of different criteria for the Male and Female groups.

\noindent\textbf{Fig.~\ref{fig:multisa}:} In a scenario involving two sensitive attributes, namely \textit{sex} (male, female) and \textit{race} (White, Black), demographic parity is achieved concerning \textit{sex} but not \textit{race}.

\noindent\textbf{Fig.~\ref{fig:prisma}:} PRISMA diagram for this review. * denotes that six studies have been overcounted due to their involvement in research across multiple directions FE: Fairness Evaluation; UM: Unfairness Mitigation; Pre: Pre-processing; In: In-processing; Post: Post-processing.

\noindent\textbf{Fig.~\ref{fig:barplot}:} Summary of data extracted from studies in our systematic review: (a) Annual trends in research on fairness in MedIA. (b) Prevalence of various medical imaging modalities, research tasks, and associated sensitive attributes.

\noindent\textbf{Fig.~\ref{fig:visual}:} Visual disparities between images with different sensitive attributes. (a)(b): images with dark skin and light skin from Fitzpatrick-17 Dataset~\citep{groh2021evaluating}; (c)(d): images of a male patient and a male patient from FairSeg Dataset~\citep{tian2024fairseg}.

\noindent\textbf{Fig.~\ref{fig:schematic diagram}:} Schematic diagram of unfairness mitigation algorithms. (a) Pre-processing methods. $D_1, D_2$: Two independent datasets; $D_{\text{origin}}, D_{\text{synthesis}}$: The original dataset and synthesized dataset. (b) In-processing methods. $\mathbb{FE}, \mathbb{TB}, \mathbb{SB}$: Feature Extractor, Target Branch, and Sensitive Branch, which are three parts of an adversarial network; $f$: latent feature vector; $y, a,\hat{y},\hat{a}$: the ground truth target task label, sensitive attributes, and their corresponding predictions generated by the neural network; $\mathcal{L}_{CE}$: Cross-Entropy loss, measuring the difference between the predicted label and the ground truth label; $\mathcal{L}_{\text{Dis}}$: Disentanglement loss, for example, MMD-Loss~\citep{borgwardt2006integrating}, measuring the distance between two distribution; $D_{00}^{10}$: requiring the maximum distance between $f_0^0$ and $f_0^1$; $d_{00}^{01}$: requiring the minimum distance between $f_0^0$ and $f_1^0$. (c) Post-processing methods. $\mathbb{NN}^*$: a pre-trained and fixed Neural Network; $y_{\text{logits}}$: predicted probability of $y$, range from 0 to 1; $\mathbb{NN}^p$: pruned $\mathbb{NN^*}$; $\Delta(\text{ACC}_0, \text{ACC}_1)$: difference between accuracy on subgroup test set $D_{a=0}$ and $D_{a=1}$.

\noindent\textbf{Fig.~\ref{fig:sources}:} Sources of unfairness and potential solutions. From top to bottom: skewed data distribution $\rightarrow$ aggregate data from multiple datasets; anatomy difference between subgroups $\rightarrow$; annotation differences for each subgroup $\rightarrow$ using causal image synthesis methods to transfer the input to the synthesis ones with opposite attribute; annotation noise $\rightarrow$involve multi-annotators to stabilize annotation; ERM-based model selection which chooses models with the highest overall performance $\rightarrow$ DTO-based model selection which consider both performance and fairness; spurious correlations between sensitive attributes and diagnosis $\rightarrow$ removing the effects of the confounder; inherited bias from the pre-train dataset $\rightarrow$ pruning the pre-trained with fairness constraints; domain gaps between the source dataset and target dataset $\rightarrow$ using domain adaptation methods to transfer models.



\end{document}

%% file: tablefc.tex
\begin{table*}[ht]
    \caption{\revise{Widely Used Criteria for Fairness}}\label{tab.fc}
    \centering
    \resizebox{\textwidth}{!}{
        \begin{threeparttable}
            \begin{tabular}{lll}
                \toprule
                    \textbf{Metrics} & \textbf{Formula}\tnote{1} & \textbf{Explanation}\tnote{2} \\
                \midrule
                Demographic Parity (DP)~\citep{dwork2012fairness} & $P(\hat{Y} = 1\mid A=0) = P(\hat{Y} = 1\mid A=1)$ & The model outcome should not be affected by any sensitive attribute. \\
                Accuracy Parity (AP)~\citep{zafar2017fairness} & $P(\hat{Y} = Y\mid A=0) = P(\hat{Y} = Y\mid A=1)$ & The model should have an equal accuracy among subgroups. \\
                Equalized Odds (EqOdds)~\citep{hardt2016equality} & $P(\hat{Y}=1 \mid A=0, Y=y) = P(\hat{Y}=1 \mid A=1, Y=y), y\in\{0, 1\}$ & The model should have an equal TPR  and FPR among subgroups.  \\
                Equal Opportunity (EqOpp)~\citep{hardt2016equality} & $P(\hat{Y}=1 \mid A=0, Y=1) = P(\hat{Y}=1 \mid A=1, Y=1)$ & The model should have an equal TPR among subgroups. \\
                \bottomrule
            \end{tabular}
            \begin{tablenotes}
                \item[1] $Y, \hat{Y} \in \{0,1\}$ denotes the ground truth label and model prediction, respectively. $A \in \{0,1\}$ denotes the sensitive attribute. Note that $Y$ and $A$ can be easily extended to multi-class situations.
                \item[2] TPR\@: True Positive Rate; FPR\@: False Positive Rate.
            \end{tablenotes}
        \end{threeparttable}
    }
\end{table*}

%% file: tableed.tex
\begin{table*}[htbp]
    \caption{\revise{Overview of Studies in Fair MedIA}}\label{tab:extract_data}
    \centering
    \resizebox{\textwidth}{!}{
        \begin{threeparttable}
            \begin{tabular}{clllllll}
            \toprule
            \textbf{Research Area}\tnote{1}	&	    \textbf{Year}	&	\textbf{Citation}	&	\textbf{Imaging Modality}	&	\textbf{Dataset}	&	\textbf{Task}\tnote{2}	&	\textbf{Sensitive Attributes}	&	\textbf{Fairness Metrics}\tnote{3}	\\
            \midrule
            \multirow{19}{*}{FE-Benchmarking}	&	2020	&	 \citep{larrazabal2020gender} 	&	Chest X-ray	&	NIH Chest-XRay14, CheXpert	&	C	&	Sex	&	$\Delta$AUC	\\
            &	2020	&	 \citep{seyyed2020chexclusion} 	&	Chest X-ray	&	MIMIC-CXR, ChestXray8, CheXpert	&	C	&	Age, Race, Sex, etc.	&	$\Delta$AUC, $\Delta$TPR	\\
            &	2020	&	 \citep{kinyanjui2020fairness} 	&	Dermatology	&	ISIC, SD-198	&	C	&	Skin tone	&	AP	\\
            &	2021	&	 \citep{pmlr-v174-zhang22a} 	&	Chest X-ray	&	 CheXpert, MIMIC-CXR  	&	 C 	&	 Race, Sex	&	$\Delta$AUC, $\Delta$BCE, $\Delta$ECE, $\Delta$TPR, $\Delta$TNR	\\
            &	2021	&	 \citep{seyyed2021underdiagnosis} 	&	Chest X-ray	&	MIMIC-CXR, CheXpert, ChestXray14	&	C	&	Age, Race, Sex	&	$\Delta$FPR, $\Delta$FNR	\\
            &	2022	&	 \citep{zong2022medfair} 	&	Multiple	&	 Multiple datasets 	&	 C 	&	 Age, Race, Sex, Skin tone	&	Max-Min Fairness,$\Delta$AUC, $\Delta$BCE, $\Delta$ECE, $\Delta$TPR, $\Delta$FPR, $\Delta$FNR, EqOdds	\\
            &	2022	&	 \citep{ioannou2022study} 	&	Brain MRI	&	ADNI	&	S	&	Race, Sex	&	$\Delta${DSC}	\\
            &	2022	&	 \citep{petersen2022feature} 	&	Brain MRI	&	ADNI	&	C	&	Sex	&	$\Delta$AUC, $\Delta$ACC	\\
            &	2022	&	 \citep{puyol2022fairness} 	&	Heart MRI	&	UK Biobank	&	S	&	Race, Sex	&	$\Delta${DSC}	\\
            &	2022	&	 \citep{stanley2022fairness} 	&	Brain MRI	&	ABCD	&	C	&	Race	&	AP	\\
            &	2022	&	 \citep{ribeiro2022fair} 	&	Brain MRI	&	ABIDE	&	C	&	Sex,Sites	&	$\Delta$AUC, $\Delta$TPR	\\
            &	2023	&	 \citep{lee_investigation_2023} 	&	Brain MRI	&	UK BioBank	&	S	&	Race, Sex	&	$\Delta${DSC}, Performance Range, SER-{DSC}, STD-{DSC}, Bias Trend	\\
            &	2023	&	 \citep{kalb_2023_revisiting} 	&	Dermatology	&	ISIC	&	C	&	Skin tone	&	$\Delta$BACC	\\
            &	2023	&	 \citep{glocker2021algorithmic} 	&	Chest X-ray	&	CheXpert, MIMIC-CXR	&	C	&	Race, Sex	&	$\Delta$TPR, $\Delta$AUC	\\
            &	2023	&	\citep{yuan_algorithmic_2023}	&	Brain MRI	&	ADNI	&	C	&	Sex, Race	&	DP, EqOpp, EqOdds	\\
            &	2023	&	\citep{huti_investigation_2023}	&	Breast MRI	&	TCIA	&	C	&	Race	&	AP	\\
            &	2023	&	\citep{wesarg_auditing_2023}	&	Brain MRI	&	ADNI	&	C	&	Age, Race, Sex	&	$\Delta$ECE, $\Delta$FPR, $\Delta$FNR	\\
            &	2023	&	\citep{salahuddin_head_2023}	&	PET/CT	&	HECKTOR	&	S	&	Age, Sex	&	$\Delta${DSC}, $\Delta$TPR	\\
            &	2023	&	\citep{klingenberg_higher_2023}	&	Brain MRI	&	ADNI	&	C	&	Sex	&	$\Delta$BACC	\\
            \midrule													
            \multirow{7}{*}{FE-Discovery}	&	2021	&	 \citep{lu2021evaluating} 	&	Mammography	&	 DMIST 	&	 C 	&	 Race	&	$\Delta$AUC	\\
            &	2023	&	 \citep{liang2023visualizing} 	&	Chest X-ray	&	 CheXpert 	&	 / 	&	 Race	&		\\
            &	2023	&	 \citep{jones2023role} 	&	Multiple	&	 Multiple datasets 	&	 C 	&	 Age, Race, Sex, Skin tone	&	$\Delta$AUC, $\Delta$ACC	\\
            &	2023	&	\citep{weng_are_2023}	&	Chest X-ray	&	NIH ChestX-ray8, CheXpert	&	C	&	Sex	&	$\Delta$AUC	\\
            &	2023	&	\citep{bercea_bias_2023}	&	Brain MRI	&	ADNI	&	AD	&	Race, Sex	&	$\Delta$MAE	\\
            &	2023	&	\citep{jimenez-sanchez_detecting_2023}	&	Chest X-ray	&	CheXpert, NIH ChestX-ray 14	&	C	&	Sex	&	$\Delta$AUC	\\
            &	2023	&	 \citep{mehta2023evaluating} 	&	Multiple	&	 ISIC, BraTS, ADAS13 	&	 C, S, R 	&	 Age, Sex	&	Ap, $\Delta${DSC}, $\Delta$RMSE	\\
            \midrule																
            \multirow{2}{*}{UM-Pre-Re-distribution}	&	2021	&	 \citep{puyol2021fairness} 	&	Heart MRI	&	 UK BioBank 	&	 S 	&	 Race, Sex 	&	$\Delta${DSC}, SER-{DSC}, STD-{DSC}	\\
            &	2023	&	 \citep{oguguo2023comparative} 	&	Chest X-ray, Dermatology	&	 JSRT, Stanford DDI 	&	 C, S 	&	Age, Skin tone	&	$\Delta$IOU, $\Delta$ACC	\\
            \midrule																
            \multirow{6}{*}{UM-Pre-Harmonization}	&	2019	&	 \citep{bissoto2019constructing} 	&	Dermatology	&	 ISIC 	&	 C 	&	 Skin tone 	&	$\Delta$AUC	\\
            &	2020	&	 \citep{bissoto2020debiasing} 	&	Dermatology	&	 ISIC 	&	 C 	&	 Skin tone 	&	$\Delta$AUC	\\
            &	2021	&	 \citep{wachinger2021detect} 	&	Brain MRI	&	 Multi-site MRI 	&	 C 	&	 Sites 	&	AP	\\
            &	2022	&	 \citep{yao2022improving} 	&	Dermatology	&	 ISIC 	&	 C 	&	 Age, Sex 	&	DP, EqOpp, EqOdds	\\
            &	2023	&	 \citep{wesarg_-identification_2023} 	&	Retinal	&	 BRSET, Diabetes Center 	&	 C 	&	 Sex 	&	$\Delta$F1	\\
            &	2023	&	 \citep{yuan_edgemixup_2022} 	&	Dermatology	&	 Private 	&	 C,S 	&	 Skin tone 	&	$\Delta$AUC, $\Delta$ACC	\\
            \midrule																
            \multirow{3}{*}{UM-Pre-Aggregation}	&	2020	&	 \citep{seyyed2020chexclusion} 	&	Chest X-ray	&	MIMIC-CXR, ChestXray8, CheXpert	&	C	&	Age, Race, Sex, etc.	&	$\Delta$AUC, $\Delta$TPR	\\
            &	2021	&	 \citep{zhou2021radfusion} 	&	Chest CT	&	 CTPA 	&	 C 	&	 Age, Race, Sex 	&	$\Delta$ACC, $\Delta$AUC, $\Delta$TPR, $\Delta$TNR,$\Delta$PPV, $\Delta$NPV	\\
            &	2023	&	\citep{wang_bias_2023}	&	Brain MRI	&	iSTAGING, PHENOM, ABIDE	&	C	&	 Age, Race, Sex 	&	DP, EqOpp, EqOdds	\\
            \midrule																
            \multirow{4}{*}{UM-Pre-Synthesis}	&	2020	&	 \citep{joshi2021ai} 	&	Retinal	&	 AREDS 	&	 C 	&	 Race 	&	AP	\\
            &	2021	&	 \citep{burlina2021addressing} 	&	Retinal	&	 Kaggle EyePACS 	&	 C 	&	 Skin tone	&	$\Delta$AUC, $\Delta$AUC, $\Delta$TPR, $\Delta$TNR	\\
            &	2022	&	 \citep{pakzad2022circle} 	&	Dermatology	&	 Fitzpatrick-17k 	&	 C 	&	 Skin tone 	&	EqOdds, NAR	\\
            &	2023	&	\citep{pombo_equitable_2023}	&	Brain MRI	&	OASIS, UK BioBank	&	C	&	Age, Sex	&	$\Delta$BACC, $\Delta$TPR, $\Delta$PPV	\\
            \midrule																
            \multirow{8}{*}{UM-In-Adversarial}	&	2020	&	 \citep{zhao2020training} 	&	Brain MRI, Bone CT	&	 Private dataset 	&	 C 	&	 Age, Sex	&	$\Delta$BACC, $\Delta$TPR, $\Delta$PPV	\\
            &	2020	&	 \citep{abbasi2020risk} 	&	Dermatology	&	 ISIC 	&	 C 	&	 Age, Sex	&	$\Delta$AUC	\\
            &	2021	&	 \citep{adeli2021representation} 	&	Brain MRI	&	 Private dataset 	&	 C 	&	 Age, Sex	&	Correlation	\\
            &	2021	&	 \citep{li2021estimating} 	&	Dermatology	&	 ISIC 	&	 C 	&	 Age, Sex, Skin tone 	&	DP, EqOpp, EqOdds	\\
            &	2021	&	 \citep{puyol2021fairness} 	&	Heart MRI	&	 UK BioBank 	&	 S 	&	 Race, Sex	&	$\Delta${DSC}, SER-{DSC}, STD-{DSC}	\\
            &	2022	&	 \citep{bevan2022detecting} 	&	Dermatology	&	 ISIC 	&	 C 	&	 Skin tone	&	$\Delta$AUC	\\
            &	2022	&	\citep{baxter_disproportionate_2022}	&	Brain MRI	&	ABCD	&	C	&	Race	&	AP	\\
            &	2023	&	\citep{correa_systematic_2022}	&	Chest X-ray, Mammography	&	Private	&	C	&	Race	&	$\Delta$TPR	\\
            \midrule																
            \multirow{6}{*}{UM-In-Constraints}	&	2021	&	 \citep{cherepanova2021technical} 	&	Chest X-ray	&	 CheXpert 	&	 C 	&	Age, Sex	&	$\Delta$AUC	\\
            &	2022	&	 \citep{marcinkevics2022debiasing} 	&	Chest X-ray	&	MIMIC-CXR	&	C	&	Race, Sex	&	DP, EqOdds	\\
            &	2022	&	 \citep{luo2022pseudo} 	&	Chest X-ray	&	MIMIC-CXR, ChestXray8	&	C	&	Sex, Sites	&	$\Delta$AUC	\\
            &	2022	&	\citep{gronowski_renyi_2022}	&	Retinal	&	EyePACS	&	S	&	Skin tone	&	DP, EqOdds	\\
            &	2023	&	\citep{lin_improving_2023}	&	Multiple	&	MIDRC, AREDS, OHTS, MIMIC-CXR	&	C	&	 Age, Race, Sex 	&	Relative Change	\\
            &	2023	&	 \citep{oguguo2023comparative} 	&	Chest X-ray, Dermatology	&	 JSRT, Stanford DDI 	&	 C, S 	&	Age, Skin tone	&	$\Delta$IOU, $\Delta$ACC	\\
            \midrule																
            \multirow{5}{*}{UM-In-Disentanglement}	&	2021	&	\citep{more2021confound} 	&	Brain MRI	&	HCP	&	C	&	Age	&	$\Delta$AUC, $\Delta$ACC, $\Delta$F1	\\
            &	2022	&	\citep{vento2022penalty} 	&	Brain MRI	&	USCF, ADNI, SRI	&	C	&	Age, Sex	&	Correlation	\\
            &	2022	&	\citep{lawry2022conditional} 	&	Brain MRI	&	UK BioBank, ADNI	&	C	&	Age	&	$\Delta$MSE	\\
            &	2022	&	\citep{sarhan2020fairness} 	&	Brain MRI	&	ABIDE	&	C	&	Age, Sex	&	AP	\\
            &	2023	&	\citep{deng2023fairness} 	&	Chest X-ray	&	CheXpert	&	C	&	Age, Race, Sex	&	EqOdds, {$\Delta$AUC}	\\
            \midrule																
            \multirow{2}{*}{UM-In-Contrastive}	&	2022	&	 \citep{du2022fairdisco} 	&	Dermatology	&	 Fitzpatrick-17k, Stanford DDI 	&	 C 	&	Skin tone	&	Division-ACC, DPM, EOM	\\
            &	2022	&	 \citep{pakzad2022circle}	&	Dermatology	&	 Fitzpatrick17k	&	 C 	&	Skin tone	&	EqOdds, NR-ACC	\\
            \midrule																
            \multirow{3}{*}{UM-In-MISC}	&	2021	&	\citep{fan2021fairness}	&	Dermatology	&	ISIC, Fitzpatrick17k	&	C	&	Age, Sex	&	DP, EqOdds	\\
            &	2023	&	 \citep{xu2023fairadabn} 	&	Dermatology	&	 Fitzpatrick-17k, ISIC 	&	 C 	&	 Sex, Skin tone 	&	EqOdds, NAR	\\
            &	2023	&	\citep{dutt2023fairtune}	&	Multiple	&	 Multiple datasets 	&	C	&	Age, Sex, Skin tone 	&	$\Delta$AUC, Max-Min Fairness	\\
            \midrule																
            UM-Post-Calibration	&	2023	&	 \citep{oguguo2023comparative} 	&	Chest X-ray, Dermatology	&	 JSRT, Stanford DDI 	&	 C, S 	&	Age, Skin tone	&	$\Delta$IOU, $\Delta$ACC	\\
            \midrule																
            \multirow{3}{*}{UM-Post-Purning}	&	2022	&	 \citep{marcinkevics2022debiasing} 	&	Chest X-ray	&	MIMIC-CXR	&	C	&	Race, Sex	&	DP, EqOdds	\\
            &	2022	&	 \citep{wu2022fairprune} 	&	Dermatology	&	ISIC, Fitzpatrick17k	&	C	&	Sex, Skin tone	&	EqOpp, EqOdds	\\
            &	2023	&	 \citep{wesarg_mitigating_2023} 	&	Brain MRI	&	 ADNI 	&	 C 	&	 Age, Sex	&	$\Delta$ACC, $\Delta$AUC, $\Delta$TPR, $\Delta$TNR,$\Delta$PPV, EqOdds	\\
            \bottomrule
            \end{tabular}
            \begin{tablenotes}
                \item[1] FE\@: Fairness Evaluation; UM\@: Unfairness Mitigation
                \item[2] C\@: Classification; S\@: Segmentation; R\@: Regression; AD\@: Anomaly Detection
                \item[3] TPR\@: True Positive Rate; TNR\@: True Negative Rate; FPR\@: False Positive Rate; FNR\@: False Negative Rate; PPV\@: Positive Predictive Value; NPV\@: Negative Predictive Value; AUC\@: Area Under Curve; ACC\@: Accuracy; BACC\@: Balanced Accuracy; BCE\@: Binary Cross Entropy; ECE\@: Expected Calibration Error; MAE\@: Mean Absolute Error; IOU\@: Intersection over Union;DSC\@: Dice Similarity Coefficient; RMSE\@: Root Mean Squard Error; DPM\@: Division form of DP\@; EOM:\@ Division form of EqOpp.
            \end{tablenotes}
        \end{threeparttable}
    }
\end{table*}

%% file: tabledwa.tex
\begin{table*}[htbp]
    \caption{{Available Medical Datasets for Fairness Assessment}}\label{dataset_with_attrs}
    \centering
    \resizebox{\linewidth}{!}{
        \begin{threeparttable}
            \begin{tabular}{lllccccccccr}
                \toprule
                \multirow{2}{*}{\textbf{Image Modality}} & \multirow{2}{*}{\textbf{Dataset}} & \multirow{2}{*}{\textbf{Task}\tnote{1}} & \multicolumn{8}{c}{\textbf{Sensitive Attributess}} & \multirow{2}{*}{\textbf{\# Images}} \\
                & & & \textbf{Age} & \textbf{Sex} & \textbf{Race / Skin tone} & \textbf{Marital} & \textbf{Drink} & \textbf{Smoke} & \textbf{Body Prams}\tnote{2} & \textbf{Handness} & \\
                \midrule
                \multirow{11}{*}{Chest X-ray} & CheXpert~\citep{irvin2019chexpert}& C,D & \checkmark& \checkmark& \checkmark& & & &  & & 224,316 \\
                            & NIH Chest X-ray~\citep{wang2017chestx} & C,D & \checkmark& \checkmark& & & & &  & & 112,120 \\
                            & MIMIC-CXR~\citep{johnson2019mimic} & C,D & \checkmark& \checkmark& \checkmark& \checkmark& & & & &  371,858 \\
                            & PadChest~\citep{bustos2020padchest} & C & \checkmark& \checkmark& & & & & & &  160,868\\
                            & BrixIA~\citep{borghesi2020covid,BS-Net2021} & C & \checkmark& \checkmark& & & & & & &  4,703 \\
                            & JSRT~\citep{shiraishi2000development} & C & \checkmark& \checkmark& & & & & & &  154 \\
                            & COVID-ChestXray~\citep{cohen2020covidProspective} & C,S & \checkmark& \checkmark& & & & & &  & 950 \\
                            & Montgomery County X-ray~\citep{jaeger2014two} & C,S & \checkmark& \checkmark& & & & & & &  138 \\
                            & Shenzhen Hospital X-ray~\citep{jaeger2014two} & C & \checkmark& \checkmark& & & & & & &  662 \\
                            & NIH PLCO~\citep{kramer1993national} & C & \checkmark& \checkmark& \checkmark& & & & & &  205,000 \\
                            & VinDr-CXR~\citep{nguyen2022vindr} & C & \checkmark& \checkmark& & & & & & &  18,000 \\
                \midrule
                \multirow{5}{*}{Chest CT} & NSCLC~\citep{ettinger2017non} & S & \checkmark& \checkmark& & & & & &  & 422\\
                         & COVID19-CT-dataset~\citep{shakouri2021covid19} & C,S & \checkmark& \checkmark& & & & & &  & 1,000+ \\
                         & NIH-NLST~\citep{national2011national} & D & \checkmark& \checkmark& \checkmark& \checkmark& \checkmark& \checkmark& \checkmark&  &   $\sim$15,000 \\
                         & LNDb~\citep{pedrosa2019lndb} & C,S,D & \checkmark& \checkmark& & &  & & & &  294 \\
                         & COVID-CT-MD~\citep{afshar2021covid} & C & \checkmark& \checkmark& & &  & & &  & 308 \\
                \midrule
                Kidney CT & KiTS 2019~\citep{heller2019kits19} & S,D & \checkmark& \checkmark& & &\checkmark& \checkmark& \checkmark& &  300 \\
                \midrule
                \multirow{3}{*}{Bone CT} & OAI\tnote{3} & D & \checkmark& \checkmark& \checkmark& & & & & &  26,626,000\\
                        & HNSCC-3DCT-RT~\citep{bejarano2019longitudinal} & S & \checkmark& \checkmark& & & & & &  & 31\\
                        & Digital Hand Atlas\citep{cao2003image} & L & & \checkmark& \checkmark&&&&&&1390 \\
                \midrule
                Body CT & DeepLesion~\citep{yan2018deeplesion} & D & \checkmark& \checkmark& & & & & & &  32,735\\
                \midrule
                \multirow{3}{*}{Fundus Images} & ODIR2019\tnote{4} & C & \checkmark& \checkmark& & & & & & & 5,000\\
                              & AREDS2~\citep{age1999age} & C & \checkmark& \checkmark& \checkmark& \checkmark& & & &  & 4,203 \\
                              & PAPILA~\citep{kovalyk2022papila} & C & \checkmark& \checkmark& & & & & & & 420 \\

                \midrule              
                \multirow{3}{*}{Eyes OCT} & OCTAGON~\citep{diaz2019automatic} & S & \checkmark& & & & &  & & & 233 \\
                         & JHU-OCT~\citep{he2019retinal} & S & \checkmark& & & & & & & &  35 \\
                         & OCT~\citep{farsiu2014quantitative} & C &\checkmark& & & & & & & & 384 \\
                         
                \midrule
                \multirow{1}{*}{Vessel Ultrasound} & TKTube~\citep{kwitt2013localizing}& S & \checkmark& \checkmark& \checkmark& & & & & \checkmark& 233 \\
                \midrule
                \multirow{3}{*}{Cardiac MRI} & ACDC~\citep{bernard2018deep} & C,S,D & \checkmark& & & & & & \checkmark& &  150 \\
                            & Sunnybrook~\citep{bernard2018deep} & S,D & \checkmark& \checkmark& & & & & & &  45 \\
                            & UKBioBank~\citep{petersen2015uk} & S & \checkmark& \checkmark& \checkmark& & & & & &  5,903 \\
                \midrule
                \multirow{11}{*}{Brain MRI/fMRI} & ADHD-200~\citep{adhd2012adhd} & C & \checkmark& \checkmark& & & & & & \checkmark&  200\\
                          & OASIS~\citep{marcus2007open} & C & \checkmark& \checkmark& & & & & & \checkmark&  $\sim$500\\
                          & ABIDE~\citep{di2014autism} & C & \checkmark& \checkmark& & & & & & \checkmark&  1,112\\
                          & PPMI~\citep{marek2011parkinson} & C,R & \checkmark& \checkmark& & & & & & \checkmark&  254\\
                          & BraTS 2019~\citep{menze2014multimodal} & C,S & \checkmark& & & & & & & &  335 \\
                          & CANDI~\citep{frazier2008diagnostic} & C,S,R & \checkmark& \checkmark& & & & & &  \checkmark&  103 \\
                          & Cam-CAN~\citep{shafto2014cambridge} & C & \checkmark& & & & & & & &  $\sim$650 \\
                          & FCP~\citep{milham2011international} & C & \checkmark& \checkmark& & & & & & &  /\tnote{5}\\
                          & ADNI~\citep{jack2008alzheimer} & C,D & \checkmark& \checkmark& & & & & & \checkmark&  800 \\
                          & ABCD~\citep{karcher2021abcd} & S & \checkmark& \checkmark& \checkmark& & & & & &  4,547\\
                          & UKBioBank~\citep{petersen2015uk} & S & \checkmark& \checkmark& \checkmark& & & & \checkmark& &  22,528 \\
                \midrule
                \multirow{2}{*}{Dermoscopy} & ISIC~\citep{codella2019skin,rotemberg2021patient} & C,S & \checkmark& \checkmark& & & & & & &  10,000+ \\
                           & Fitzpatrick-17k~\citep{groh2021evaluating,groh2022towards} & C & & & \checkmark& & & & & &  16,577 \\
                \bottomrule
            \end{tabular}
            \begin{tablenotes}
                \item[1] C\@: Classification, S\@: Segmentation, D\@: Detection, L\@: Landmark Detection, R\@: Registration
                \item[2] Including Weight, Height and BMI
                \item[3] https://nda.nih.gov/oai
                \item[4] https://odir2019.grand-challenge.org/
                \item[5] This dataset contains more than 1,200 fMRI datasets collected independently at 33 sites.
            \end{tablenotes}
        \end{threeparttable}
    }
\end{table*}

%% file: tableglossary.tex
\begin{table*}[ht]
    \caption{{\revise{A Glossary of Technical Terms Used in Fairness Research}}}\label{tab:glossary}
    \centering
    \resizebox{\textwidth}{!}{
    \begin{tabular}{p{0.25\textwidth}p{0.75\textwidth}}
        \toprule
            \textbf{Technical Terms} & \textbf{Definition} \\
            \midrule
            Adversarial Learning & A type of neural network consisting of a target branch and an adversarial branch, aims to predict two variables,  $\hat{Y}$ (class label) and $\hat{A}$ (sensitive attribute), respectively. However, although the loss function of the adversarial branch minimizes the prediction error of $\hat{A}$, the gradient is reversed by a gradient reversal layer~\citep{ganin2015unsupervised}, which pushes the network not to recognize $A$.  \\
            \midrule
            Disentanglement Learning & The neural network \revise{extract} high-level features which is a mixture of information from the target task and the attribute. Disentanglement learning projects the feature vector into another space, where the information is split into independent components~\citep{creager2019flexibly}.\\
            \midrule
            Contrastive Learning & Contrastive learning takes paired samples from either the same class or different classes and uses loss \revise{functions} that maximizes the disparity between features from different classes and minimizes the disparity between features from the same class.~\citep{chuang2020debiased}\\
            \midrule
            Model Calibration & Model calibration adjusts the predicted threshold of the model's output logits per subgroup, for example, \revise{regarding} output logits larger than 0.65 (or 0.53) as the positive for the Male (or Female) subgroup, to ensure that the model has equal fairness criteria in each subgroup.~\citep{oguguo2023comparative}\\
            \midrule
            Model Pruning & For a pre-trained neural network, model pruning inspects all the neurons in the network, and removes some of them based on some fairness metrics, \revise{resulting} in a network with fewer parameters.~\citep{wu2022fairprune} \\
            \midrule
            Uncertainty & A variable that measures the stability of the neural network when facing the fluctuation of the input data.~\citep{jungo}\\
         \bottomrule
    \end{tabular}
    }
\end{table*}

%% file: tablefr.tex
\begin{table*}[tbp]
    \centering
    \caption{{Comparisons between Fair MedIA and Fair FR}}\label{tab:FR_align}
    \resizebox{\textwidth}{!}{
    \begin{tabular}{lll}
        \toprule
         & \textbf{Fair MedIA} & \textbf{Fair FR} \\
        \midrule
        \textbf{Image Modality} & \makecell[l]{2D\@: X-ray~\citep{pmlr-v174-zhang22a}, dermoscopy~\citep{xu2023fairadabn}, mammography~\citep{schwartz2021association} \\ 3D\@: MRI~\citep{puyol2021fairness}, PET/CT~\citep{salahuddin_head_2023}} & RGB images~\citep{karkkainen2021fairface}\\
        \midrule
        \textbf{Amount of Samples} & Range from tens to tens of thousands & Range from thousands to millions\\
        \midrule
        \textbf{Sensitive Attributes} & \makecell[l]{Sex, Age, Race, Skin tone, etc. \\(describing demographics)} &  \makecell[l]{Sex, Age, Race, Skin tone, Hair Color, etc. \\(describing appearance)~\citep{liu2015deep}}\\ 
        \midrule
        \textbf{Dataset Composition} & Skewed attribute distribution & \makecell[l]{Skewed attribute distribution: CelebA~\citep{liu2015deep}\\ Balanced attribute distribution: FairFace~\citep{karkkainen2021fairface}}\\
         \bottomrule
    \end{tabular}
    }

\end{table*}

%% file: 0_main.bbl
\begin{thebibliography}{100}
\providecommand{\url}[1]{#1}
\csname url@samestyle\endcsname
\providecommand{\newblock}{\relax}
\providecommand{\bibinfo}[2]{#2}
\providecommand{\BIBentrySTDinterwordspacing}{\spaceskip=0pt\relax}
\providecommand{\BIBentryALTinterwordstretchfactor}{4}
\providecommand{\BIBentryALTinterwordspacing}{\spaceskip=\fontdimen2\font plus
\BIBentryALTinterwordstretchfactor\fontdimen3\font minus
  \fontdimen4\font\relax}
\providecommand{\BIBforeignlanguage}[2]{{%
\expandafter\ifx\csname l@#1\endcsname\relax
\typeout{** WARNING: IEEEtran.bst: No hyphenation pattern has been}%
\typeout{** loaded for the language `#1'. Using the pattern for}%
\typeout{** the default language instead.}%
\else
\language=\csname l@#1\endcsname
\fi
#2}}
\providecommand{\BIBdecl}{\relax}
\BIBdecl

\bibitem{stanley2022fairness}
E.~A. Stanley, M.~Wilms, P.~Mouches, and N.~D. Forkert, ``Fairness-related
  performance and explainability effects in deep learning models for brain
  image analysis,'' \emph{J. Med. Imaging}, vol.~9, no.~6, p. 061102, 2022.

\bibitem{seyyed2021underdiagnosis}
L.~Seyyed-Kalantari, H.~Zhang, M.~B. McDermott, I.~Y. Chen, and M.~Ghassemi,
  ``Underdiagnosis bias of artificial intelligence algorithms applied to chest
  radiographs in under-served patient populations,'' \emph{Nat. Med.}, vol.~27,
  no.~12, pp. 2176--2182, 2021.

\bibitem{puyol2021fairness}
E.~Puyol-Antón \emph{et~al.}, ``Fairness in cardiac {{MR}} image analysis:
  {{An}} investigation of bias due to data imbalance in deep learning based
  segmentation,'' in \emph{Int. {{Conf}}. {{Med}}. {{Image Comput}}.
  {{Comput}}.-{{Assist}}. {{Interv}}.}, 2021.

\bibitem{puyol2022fairness}
E.~Puyol-Ant{\'o}n \emph{et~al.}, ``Fairness in cardiac magnetic resonance
  imaging: assessing sex and racial bias in deep learning-based segmentation,''
  \emph{Front. Cardiovasc. Med.}, vol.~9, p. 859310, 2022.

\bibitem{seyyed2020chexclusion}
L.~Seyyed-Kalantari, G.~Liu, M.~McDermott, I.~Y. Chen, and M.~Ghassemi,
  ``{{CheXclusion}}: {{Fairness}} gaps in deep chest {{X-ray}} classifiers,''
  in \emph{Biocomput. 2021 {{Proc}}. {{Pac}}. {{Symp}}.}, 2020, pp. 232--243.

\bibitem{glocker2021algorithmic}
B.~Glocker, C.~Jones, M.~Bernhardt, and S.~Winzeck, ``Algorithmic encoding of
  protected characteristics in chest {{X-ray}} disease detection models,''
  \emph{Ebiomedicine}, vol.~89, 2023.

\bibitem{ribeiro2022fair}
F.~Ribeiro, V.~Shumovskaia, T.~Davies, and I.~Ktena, ``How fair is your graph?
  {{Exploring}} fairness concerns in neuroimaging studies,'' in \emph{Mach.
  {{Learn Heal}}. {{Conf}}}, 2022, pp. 459--478.

\bibitem{ioannou2022study}
S.~Ioannou, H.~Chockler, A.~Hammers, A.~P. King, and A.~D.~N. Initiative, ``A
  study of demographic bias in {{CNN-based}} brain {{MR}} segmentation,'' in
  \emph{Mach. {{Learn}}. {{Clin}}. {{Neuroimaging}} 5th {{Int}}. {{Workshop
  MLCN}} 2022 {{Held Conjunction MICCAI}} 2022 {{Singap}}. {{Sept}}. 18 2022
  {{Proc}}.}, 2022, pp. 13--22.

\bibitem{petersen2022feature}
E.~Petersen \emph{et~al.}, ``Feature robustness and sex differences in medical
  imaging: a case study in mri-based alzheimer’s disease detection,'' in
  \emph{International Conference on Medical Image Computing and
  Computer-Assisted Intervention}.\hskip 1em plus 0.5em minus 0.4em\relax
  Springer, 2022, pp. 88--98.

\bibitem{cherepanova2021technical}
\BIBentryALTinterwordspacing
V.~Cherepanova, V.~Nanda, M.~Goldblum, J.~P. Dickerson, and T.~Goldstein,
  ``Technical challenges for training fair neural networks,'' \emph{CoRR}, vol.
  abs/2102.06764, 2021. [Online]. Available:
  \url{https://arxiv.org/abs/2102.06764}
\BIBentrySTDinterwordspacing

\bibitem{xu2023fairadabn}
Z.~Xu, S.~Zhao, Q.~Quan, Q.~Yao, and S.~K. Zhou, ``{{FairAdaBN}}:
  {{Mitigating}} unfairness with adaptive batch normalization and its
  application to dermatological disease classification,'' in \emph{Med. {{Image
  Comput}}. {{Comput}}. {{Assist}}. {{Interv}}. – {{MICCAI}} 2023}, 2023, pp.
  307--317.

\bibitem{mehta2023evaluating}
R.~Mehta, C.~Shui, and T.~Arbel, ``Evaluating the fairness of deep learning
  uncertainty estimates in medical image analysis,'' in \emph{Med. {{Imaging
  Deep Learn}}.}, ser. Proceedings of Machine Learning Research, I.~Oguz
  \emph{et~al.}, Eds., vol. 227.\hskip 1em plus 0.5em minus 0.4em\relax {PMLR},
  2023, pp. 1453--1492.

\bibitem{larrazabal2020gender}
A.~J. Larrazabal, N.~Nieto, V.~Peterson, D.~H. Milone, and E.~Ferrante,
  ``Gender imbalance in medical imaging datasets produces biased classifiers
  for computer-aided diagnosis,'' \emph{Proc. Natl. Acad. Sci.}, vol. 117,
  no.~23, pp. 12\,592--12\,594, 2020.

\bibitem{brown2022detecting}
A.~Brown \emph{et~al.}, ``Detecting shortcut learning for fair medical {{AI}}
  using shortcut testing,'' \emph{Nat. Commun.}, vol.~14, no.~1, p. 4314, 2023.

\bibitem{adeli2021representation}
E.~Adeli \emph{et~al.}, ``Representation learning with statistical independence
  to mitigate bias,'' in \emph{Proc. {{IEEECVF Winter Conf}}. {{Appl}}.
  {{Comput}}. {{Vis}}.}, 2021, pp. 2513--2523.

\bibitem{zhang2018mitigating}
B.~H. Zhang, B.~Lemoine, and M.~Mitchell, ``Mitigating unwanted biases with
  adversarial learning,'' in \emph{Proc. 2018 {{AAAIACM Conf}}. {{AI Ethics
  Soc}}.}, 2018, pp. 335--340.

\bibitem{deng2023fairness}
W.~Deng, Y.~Zhong, Q.~Dou, and X.~Li, ``On fairness of medical image
  classification with multiple sensitive attributes via learning orthogonal
  representations,'' in \emph{Inf. {{Process}}. {{Med}}. {{Imaging}}}, 2023,
  pp. 158--169.

\bibitem{li2021estimating}
\BIBentryALTinterwordspacing
X.~Li, Z.~Cui, Y.~Wu, L.~Gu, and T.~Harada, ``Estimating and improving fairness
  with adversarial learning,'' \emph{CoRR}, vol. abs/2103.04243, 2021.
  [Online]. Available: \url{https://arxiv.org/abs/2103.04243}
\BIBentrySTDinterwordspacing

\bibitem{pakzad2022circle}
A.~Pakzad, K.~Abhishek, and G.~Hamarneh, ``{{CIRCLe}}: {{Color}} invariant
  representation learning for unbiased classification of skin lesions,'' in
  \emph{Proc. 17th {{Eur}}. {{Conf}}. {{Comput}}. {{Vis}}. {{ECCV}} - {{ISIC
  Skin Image Anal}}. {{Workshop}}}, 2022.

\bibitem{yang2023algorithmic}
J.~Yang, A.~A. Soltan, D.~W. Eyre, and D.~A. Clifton, ``Algorithmic fairness
  and bias mitigation for clinical machine learning with deep reinforcement
  learning,'' \emph{Nat. Mach. Intell.}, vol.~5, no.~8, pp. 884--894, 2023.

\bibitem{yang2023adversarial}
J.~Yang, A.~A. Soltan, D.~W. Eyre, Y.~Yang, and D.~A. Clifton, ``An adversarial
  training framework for mitigating algorithmic biases in clinical machine
  learning,'' \emph{NPJ Digit. Med.}, vol.~6, no.~1, p.~55, 2023.

\bibitem{celeste2023ethnic}
C.~Celeste \emph{et~al.}, ``Ethnic disparity in diagnosing asymptomatic
  bacterial vaginosis using machine learning,'' \emph{NPJ Digit. Med.}, vol.~6,
  no.~1, p. 211, 2023.

\bibitem{lara2023towards}
M.~A. Ricci~Lara, C.~Mosquera, E.~Ferrante, and R.~Echeveste, ``Towards
  unraveling calibration biases in medical image analysis,'' in \emph{Workshop
  {{Clin}}. {{Image-Based Proced}}.}, 2023, pp. 132--141.

\bibitem{beauchamp2001principles}
S.~Holm, ``Principles of biomedical ethics, 5th edn.'' \emph{J. Med. Ethics},
  vol.~28, no.~5, pp. 332--332, 2002.

\bibitem{liu2023translational}
M.~Liu \emph{et~al.}, ``A translational perspective towards clinical {{AI}}
  fairness,'' \emph{NPJ Digit. Med.}, vol.~6, no.~1, p. 172, 2023.

\bibitem{srivastava2019mathematical}
M.~Srivastava, H.~Heidari, and A.~Krause, ``Mathematical notions vs. human
  perception of fairness: {{A}} descriptive approach to fairness for machine
  learning,'' in \emph{Proc. 25th {{ACM SIGKDD Int}}. {{Conf}}. {{Knowl}}.
  {{Discov}}. {{Data Min}}.}, 2019, pp. 2459--2468.

\bibitem{jones2010building}
N.~Jones \emph{et~al.}, ``Building understanding of fairness, equality and good
  relations,'' \emph{Equal. Hum. Rights Comm. Res. Rep.}, vol.~53, 2010.

\bibitem{green2003unequal}
C.~R. Green \emph{et~al.}, ``The unequal burden of pain: {{Confronting}} racial
  and ethnic disparities in pain,'' \emph{Pain Med.}, vol.~4, no.~3, pp.
  277--294, 2003.

\bibitem{anderson2009racial}
K.~O. Anderson, C.~R. Green, and R.~Payne, ``Racial and ethnic disparities in
  pain: {{Causes}} and consequences of unequal care,'' \emph{J. Pain}, vol.~10,
  no.~12, pp. 1187--1204, 2009.

\bibitem{mittermaier2023bias}
M.~Mittermaier, M.~M. Raza, and J.~C. Kvedar, ``Bias in {{AI-based}} models for
  medical applications: {{Challenges}} and mitigation strategies,'' \emph{NPJ
  Digit. Med.}, vol.~6, no.~1, p. 113, 2023.

\bibitem{currie2021ethical}
G.~Currie and K.~E. Hawk, ``Ethical and legal challenges of artificial
  intelligence in nuclear medicine,'' in \emph{Semin. {{Nucl}}. {{Med}}.},
  vol.~51, no.~2, 2021, pp. 120--125.

\bibitem{batra2023new}
A.~M. Batra and A.~Reche, ``A new era of dental care: {{Harnessing}} artificial
  intelligence for better diagnosis and treatment,'' \emph{Cureus}, vol.~15,
  no.~11, 2023.

\bibitem{dwork2012fairness}
C.~Dwork, M.~Hardt, T.~Pitassi, O.~Reingold, and R.~Zemel, ``Fairness through
  awareness,'' in \emph{Proc. 3rd {{Innov}}. {{Theor}}. {{Comput}}. {{Sci}}.
  {{Conf}}.}, 2012, pp. 214--226.

\bibitem{zafar2017fairness}
M.~B. Zafar, I.~Valera, M.~G. Rogriguez, and K.~P. Gummadi, ``Fairness
  constraints: {{Mechanisms}} for fair classification,'' in \emph{Artif.
  {{Intell}}. {{Stat}}.}, 2017, pp. 962--970.

\bibitem{hardt2016equality}
M.~Hardt, E.~Price, and N.~Srebro, ``Equality of opportunity in supervised
  learning,'' \emph{Adv. Neural Inf. Process. Syst.}, vol.~29, 2016.

\bibitem{barocas2017fairness}
S.~Barocas, M.~Hardt, and A.~Narayanan, ``Fairness in machine learning,''
  \emph{Nips Tutor.}, vol.~1, p.~2, 2017.

\bibitem{lahoti2020fairness}
P.~Lahoti \emph{et~al.}, ``Fairness without demographics through adversarially
  reweighted learning,'' \emph{Adv. Neural Inf. Process. Syst.}, vol.~33, pp.
  728--740, 2020.

\bibitem{kusner2017counterfactual}
M.~J. Kusner, J.~Loftus, C.~Russell, and R.~Silva, ``Counterfactual fairness,''
  \emph{Adv. Neural Inf. Process. Syst.}, vol.~30, 2017.

\bibitem{mbakwe2023fairness}
A.~B. Mbakwe, I.~Lourentzou, L.~A. Celi, and J.~T. Wu, ``Fairness metrics for
  health {{AI}}: {{We}} have a long way to go,'' \emph{Ebiomedicine}, vol.~90,
  2023.

\bibitem{bird2020fairlearn}
S.~Bird \emph{et~al.}, ``Fairlearn: {{A}} toolkit for assessing and improving
  fairness in {{AI}},'' \emph{Microsoft Tech Rep MSR-TR-2020-32}, 2020.

\bibitem{wang2020mitigating}
M.~Wang and W.~Deng, ``Mitigating bias in face recognition using skewness-aware
  reinforcement learning,'' in \emph{Proc. {{IEEECVF Conf}}. {{Comput}}.
  {{Vis}}. {{Pattern Recognit}}.}, 2020, pp. 9322--9331.

\bibitem{kinyanjui2020fairness}
N.~M. Kinyanjui \emph{et~al.}, ``Fairness of classifiers across skin tones in
  dermatology,'' in \emph{Int. {{Conf}}. {{Med}}. {{Image Comput}}.
  {{Comput}}.-{{Assist}}. {{Interv}}.}, 2020, pp. 320--329.

\bibitem{pmlr-v174-zhang22a}
H.~Zhang \emph{et~al.}, ``Improving the fairness of chest x-ray classifiers,''
  in \emph{Proc. {{Conf}}. {{Health Inference Learn}}.}, vol. 174,
  2022-04-07/2022-04-08, pp. 204--233.

\bibitem{zong2022medfair}
Y.~Zong, Y.~Yang, and T.~Hospedales, ``{{MEDFAIR}}: {{Benchmarking}} fairness
  for medical imaging,'' in \emph{Int. {{Conf}}. {{Learn}}. {{Represent}}.
  {{ICLR}}}, 2023.

\bibitem{lee_investigation_2023}
T.~Lee \emph{et~al.}, ``An investigation into the impact of deep learning model
  choice on sex and race bias in cardiac {{MR}} segmentation,'' in \emph{Clin.
  {{Image-Based Proced}}. {{Fairness AI Med}}. {{Imaging Ethical Philos}}.
  {{Issues Med}}. {{Imaging}}}, 2023, pp. 215--224.

\bibitem{kalb_2023_revisiting}
T.~Kalb \emph{et~al.}, ``Revisiting skin tone fairness in dermatological lesion
  classification,'' in \emph{Clin. {{Image-Based Proced}}. {{Fairness AI Med}}.
  {{Imaging Ethical Philos}}. {{Issues Med}}. {{Imaging}}}, 2023, pp. 246--255.

\bibitem{yuan_algorithmic_2023}
C.~Yuan, K.~A. Linn, and R.~A. Hubbard, ``Algorithmic {{Fairness}} of {{Machine
  Learning Models}} for {{Alzheimer Disease Progression}},'' \emph{JAMA Netw.
  Open}, vol.~6, no.~11, p. e2342203, 2023-11.

\bibitem{huti_investigation_2023}
M.~Huti, T.~Lee, E.~Sawyer, and A.~P. King, ``An {{Investigation}} into {{Race
  Bias}} in {{Random Forest Models Based}} on {{Breast DCE-MRI Derived
  Radiomics Features}},'' \emph{Lect. Notes Comput. Sci. Subser. Lect. Notes
  Artif. Intell. Lect. Notes Bioinforma.}, vol. 14242 LNCS, pp. 225--234, 2023.

\bibitem{wesarg_auditing_2023}
V.~N. Dang \emph{et~al.}, ``Auditing {{Unfair Biases}} in {{CNN-Based
  Diagnosis}} of {{Alzheimer}}’s {{Disease}},'' in \emph{Clinical
  {{Image-Based Procedures}}, {{Fairness}} of {{AI}} in {{Medical Imaging}},
  and {{Ethical}} and {{Philosophical Issues}} in {{Medical Imaging}}}, 2023,
  vol. 14242, pp. 172--182.

\bibitem{salahuddin_head_2023}
Z.~Salahuddin \emph{et~al.}, ``From {{Head}} and {{Neck Tumour}} and {{Lymph
  Node Segmentation}} to {{Survival Prediction}} on {{PET}}/{{CT}}: {{An
  End-to-End Framework Featuring Uncertainty}}, {{Fairness}}, and
  {{Multi-Region Multi-Modal Radiomics}}.'' \emph{Cancers}, vol.~15, no.~7,
  2023-03.

\bibitem{klingenberg_higher_2023}
M.~Klingenberg \emph{et~al.}, ``Higher performance for women than men in
  {{MRI-based Alzheimer}}’s disease detection,'' \emph{Alzheimers Res.
  Ther.}, vol.~15, no.~1, p.~84, 2023.

\bibitem{lu2021evaluating}
\BIBentryALTinterwordspacing
C.~Lu, A.~Lemay, K.~Hoebel, and J.~Kalpathy{-}Cramer, ``Evaluating subgroup
  disparity using epistemic uncertainty in mammography,'' \emph{CoRR}, vol.
  abs/2107.02716, 2021. [Online]. Available:
  \url{https://arxiv.org/abs/2107.02716}
\BIBentrySTDinterwordspacing

\bibitem{liang2023visualizing}
H.~Liang, K.~Ni, and G.~Balakrishnan, ``Visualizing chest {{X-ray}} dataset
  biases using {{GANs}},'' in \emph{Med. {{Imaging Deep Learn}}. {{Short Pap}}.
  {{Track}}}, 2023.

\bibitem{jones2023role}
C.~Jones, M.~Roschewitz, and B.~Glocker, ``The {{Role}} of {{Subgroup
  Separability}} in {{Group-Fair Medical Image Classification}},'' in
  \emph{Med. {{Image Comput}}. {{Comput}}. {{Assist}}. {{Interv}}. –
  {{MICCAI}} 2023}, 2023.

\bibitem{weng_are_2023}
N.~Weng, S.~Bigdeli, E.~Petersen, and A.~Feragen, ``Are sex-based physiological
  differences the cause of gender bias for chest x-{{Ray}} diagnosis?'' in
  \emph{Clin. {{Image-Based Proced}}. {{Fairness AI Med}}. {{Imaging Ethical
  Philos}}. {{Issues Med}}. {{Imaging}}}, 2023, pp. 142--152.

\bibitem{bercea_bias_2023}
C.~I. Bercea \emph{et~al.}, ``Bias in unsupervised anomaly detection in brain
  {{MRI}},'' in \emph{Clin. {{Image-Based Proced}}. {{Fairness AI Med}}.
  {{Imaging Ethical Philos}}. {{Issues Med}}. {{Imaging}}}, 2023, pp. 122--131.

\bibitem{jimenez-sanchez_detecting_2023}
A.~Jimenez-Sanchez, D.~Juodelyte, B.~Chamberlain, and V.~Cheplygina,
  ``Detecting {{Shortcuts}} in {{Medical Images}} - {{A Case Study}} in {{Chest
  X-Rays}},'' in \emph{Proc. - {{Int}}. {{Symp}}. {{Biomed}}. {{Imaging}}},
  vol. 2023-April, 2023.

\bibitem{oguguo2023comparative}
T.~Oguguo \emph{et~al.}, ``A comparative study of fairness in medical machine
  learning,'' in \emph{2023 {{IEEE}} 20th {{Int}}. {{Symp}}. {{Biomed}}.
  {{Imaging ISBI}}}, 2023, pp. 1--5.

\bibitem{bissoto2019constructing}
A.~Bissoto, M.~Fornaciali, E.~Valle, and S.~Avila, ``({{De}}) constructing bias
  on skin lesion datasets,'' in \emph{Proc. {{IEEECVF Conf}}. {{Comput}}.
  {{Vis}}. {{Pattern Recognit}}. {{Workshop}}}, 2019, pp. 0--0.

\bibitem{bissoto2020debiasing}
A.~Bissoto, E.~Valle, and S.~Avila, ``Debiasing skin lesion datasets and
  models? {{Not}} so fast,'' in \emph{Proc. {{IEEECVF Conf}}. {{Comput}}.
  {{Vis}}. {{Pattern Recognit}}. {{Workshop}}}, 2020, pp. 740--741.

\bibitem{wachinger2021detect}
C.~Wachinger, A.~Rieckmann, S.~Pölsterl, A.~D.~N. Initiative \emph{et~al.},
  ``Detect and correct bias in multi-site neuroimaging datasets,'' \emph{Med.
  Image Anal.}, vol.~67, p. 101879, 2021.

\bibitem{yao2022improving}
R.~Yao, Z.~Cui, X.~Li, and L.~Gu, ``Improving fairness in image classification
  via sketching,'' in \emph{Workshop {{Trust}}. {{Socially Responsible Mach}}.
  {{Learn}}. {{NeurIPS}} 2022}, 2022.

\bibitem{wesarg_-identification_2023}
C.~Wu \emph{et~al.}, ``De-identification and {{Obfuscation}} of {{Gender
  Attributes}} from {{Retinal Scans}},'' in \emph{Clinical {{Image-Based
  Procedures}}, {{Fairness}} of {{AI}} in {{Medical Imaging}}, and {{Ethical}}
  and {{Philosophical Issues}} in {{Medical Imaging}}}, 2023, vol. 14242, pp.
  91--101.

\bibitem{yuan_edgemixup_2022}
H.~Yuan \emph{et~al.}, ``{{EdgeMixup}}: {{Embarrassingly}} simple data
  alteration to improve lyme disease lesion segmentation and diagnosis
  fairness,'' in \emph{Med. {{Image Comput}}. {{Comput}}. {{Assist}}.
  {{Interv}}. – {{MICCAI}} 2023}, 2023, pp. 374--384.

\bibitem{zhou2021radfusion}
\BIBentryALTinterwordspacing
Y.~Zhou \emph{et~al.}, ``Radfusion: Benchmarking performance and fairness for
  multimodal pulmonary embolism detection from {CT} and {EHR},'' \emph{CoRR},
  vol. abs/2111.11665, 2021. [Online]. Available:
  \url{https://arxiv.org/abs/2111.11665}
\BIBentrySTDinterwordspacing

\bibitem{wang_bias_2023}
R.~Wang, P.~Chaudhari, and C.~Davatzikos, ``Bias in machine learning models can
  be significantly mitigated by careful training: {{Evidence}} from
  neuroimaging studies.'' \emph{Proc. Natl. Acad. Sci. U. S. A.}, vol. 120,
  no.~6, p. e2211613120, 2023-02.

\bibitem{joshi2021ai}
\BIBentryALTinterwordspacing
N.~Joshi and P.~M. Burlina, ``{AI} fairness via domain adaptation,''
  \emph{CoRR}, vol. abs/2104.01109, 2021. [Online]. Available:
  \url{https://arxiv.org/abs/2104.01109}
\BIBentrySTDinterwordspacing

\bibitem{burlina2021addressing}
P.~Burlina, N.~Joshi, W.~Paul, K.~D. Pacheco, and N.~M. Bressler, ``Addressing
  artificial intelligence bias in retinal diagnostics,'' \emph{Transl. Vis.
  Sci. Technol.}, vol.~10, no.~2, pp. 13--13, 2021.

\bibitem{pombo_equitable_2023}
G.~Pombo \emph{et~al.}, ``Equitable modelling of brain imaging by
  counterfactual augmentation with morphologically constrained {{3D}} deep
  generative models.'' \emph{Med. Image Anal.}, vol.~84, p. 102723, 2023-02.

\bibitem{zhao2020training}
Q.~Zhao, E.~Adeli, and K.~M. Pohl, ``Training confounder-free deep learning
  models for medical applications,'' \emph{Nat. Commun.}, vol.~11, no.~1, pp.
  1--9, 2020.

\bibitem{abbasi2020risk}
S.~Abbasi-Sureshjani, R.~Raumanns, B.~E. Michels, G.~Schouten, and
  V.~Cheplygina, ``Risk of training diagnostic algorithms on data with
  demographic bias,'' in \emph{Interpretable and Annotation-Efficient Learning
  for Medical Image Computing}, 2020, pp. 183--192.

\bibitem{bevan2022detecting}
P.~J. Bevan and A.~Atapour-Abarghouei, ``Detecting melanoma fairly: {{Skin}}
  tone detection and debiasing for skin lesion classification,'' in
  \emph{{{MICCAI Workshop Domain Adapt}}. {{Represent}}. {{Transf}}.}, 2022,
  pp. 1--11.

\bibitem{baxter_disproportionate_2022}
E.~A.~M. Stanley, M.~Wilms, and N.~D. Forkert, ``Disproportionate {{Subgroup
  Impacts}} and {{Other Challenges}} of {{Fairness}} in {{Artificial
  Intelligence}} for {{Medical Image Analysis}},'' in \emph{Ethical and
  {{Philosophical Issues}} in {{Medical Imaging}}, {{Multimodal Learning}} and
  {{Fusion Across Scales}} for {{Clinical Decision Support}}, and {{Topological
  Data Analysis}} for {{Biomedical Imaging}}}, 2022, vol. 13755, pp. 14--25.

\bibitem{correa_systematic_2022}
R.~Correa \emph{et~al.}, ``A {{Systematic Review}} of ‘{{Fair}}’ {{AI Model
  Development}} for {{Image Classification}} and {{Prediction}},'' \emph{J.
  Med. Biol. Eng.}, vol.~42, no.~6, pp. 816--827, 2022.

\bibitem{marcinkevics2022debiasing}
R.~Marcinkevics, E.~Ozkan, and J.~E. Vogt, ``Debiasing deep chest x-ray
  classifiers using intra-and post-processing methods,'' in \emph{Mach {{Learn
  Heal}}. {{Conf}}}, 2022, pp. 504--536.

\bibitem{luo2022pseudo}
L.~Luo, D.~Xu, H.~Chen, T.-T. Wong, and P.-A. Heng, ``Pseudo bias-balanced
  learning for debiased chest x-{{Ray}} classification,'' in \emph{Med. {{Image
  Comput}}. {{Comput}}. {{Assist}}. {{Interv}}. – {{MICCAI}} 2022}, 2022, pp.
  621--631.

\bibitem{gronowski_renyi_2022}
A.~Gronowski, W.~Paul, F.~Alajaji, B.~Gharesifard, and P.~Burlina, ``Rényi
  fair information bottleneck for image classification,'' in \emph{2022 17th
  {{Can}}. {{Workshop Inf}}. {{Theory CWIT}}}, 2022, pp. 11--15.

\bibitem{lin_improving_2023}
M.~Lin \emph{et~al.}, ``Improving model fairness in image-based computer-aided
  diagnosis.'' \emph{Nat. Commun.}, vol.~14, no.~1, p. 6261, 2023-10.

\bibitem{more2021confound}
S.~More, S.~B. Eickhoff, J.~Caspers, and K.~R. Patil, ``Confound removal and
  normalization in practice: {{A}} neuroimaging based sex prediction case
  study,'' in \emph{Mach. {{Learn}}. {{Knowl}}. {{Discov}}. {{Databases Appl}}.
  {{Data Sci}}. {{Demo Track Eur}}. {{Conf}}. {{ECML PKDD}} 2020 {{Ghent
  Belg}}. {{Sept}}. 14–18 2020 {{Proc}}. {{Part V}}}, 2021, pp. 3--18.

\bibitem{vento2022penalty}
A.~Vento, Q.~Zhao, R.~Paul, K.~M. Pohl, and E.~Adeli, ``A penalty approach for
  normalizing feature distributions to build confounder-free models,'' in
  \emph{Int. {{Conf}}. {{Med}}. {{Image Comput}}. {{Comput}}.-{{Assist}}.
  {{Interv}}.}, 2022, pp. 387--397.

\bibitem{lawry2022conditional}
A.~Lawry~Aguila, J.~Chapman, M.~Janahi, and A.~Altmann, ``Conditional vaes for
  confound removal and normative modelling of neurodegenerative diseases,'' in
  \emph{Int. {{Conf}}. {{Med}}. {{Image Comput}}. {{Comput}}.-{{Assist}}.
  {{Interv}}.}, 2022, pp. 430--440.

\bibitem{sarhan2020fairness}
M.~H. Sarhan, N.~Navab, A.~Eslami, and S.~Albarqouni, ``Fairness by learning
  orthogonal disentangled representations,'' in \emph{Comput.
  {{Vision}}–{{ECCV}} 2020 16th {{Eur}}. {{Conf}}. {{Glasg}}. {{UK August}}
  23–28 2020 {{Proc}}. {{Part XXIX}} 16}, 2020, pp. 746--761.

\bibitem{du2022fairdisco}
S.~Du, B.~Hers, N.~Bayasi, G.~Harmaneh, and R.~Garbi, ``{{FairDisCo}}: {{Fairer
  AI}} in dermatology via disentanglement contrastive learning,'' in
  \emph{Proc. 17th {{Eur}}. {{Conf}}. {{Comput}}. {{Vis}}. {{Workshop ECCVW}}
  2022}, 2022.

\bibitem{fan2021fairness}
D.~Fan, Y.~Wu, and X.~Li, ``On the fairness of swarm learning in skin lesion
  classification,'' in \emph{Clinical Image-Based Procedures, Distributed and
  Collaborative Learning, Artificial Intelligence for Combating {{COVID-19}}
  and Secure and Privacy-Preserving Machine Learning}, 2021, pp. 120--129.

\bibitem{dutt2023fairtune}
R.~Dutt, O.~Bohdal, S.~A. Tsaftaris, and T.~Hospedales, ``{{FairTune}}:
  {{Optimizing}} parameter efficient fine tuning for fairness in medical image
  analysis,'' in \emph{Twelfth {{Int}}. {{Conf}}. {{Learn}}. {{Represent}}.},
  2024.

\bibitem{wu2022fairprune}
Y.~Wu, D.~Zeng, X.~Xu, Y.~Shi, and J.~Hu, ``{{FairPrune}}: {{Achieving}}
  fairness through pruning for dermatological disease diagnosis,'' in
  \emph{Med. {{Image Comput}}. {{Comput}}. {{Assist}}. {{Interv}}. –
  {{MICCAI}} 2022}, 2022, pp. 743--753.

\bibitem{wesarg_mitigating_2023}
Y.-Y. Huang, V.~Chiuwanara, C.-H. Lin, and P.-C. Kuo, ``Mitigating {{Bias}} in
  {{MRI-Based Alzheimer}}’s {{Disease Classifiers Through Pruning}} of {{Deep
  Neural Networks}},'' in \emph{Clinical {{Image-Based Procedures}},
  {{Fairness}} of {{AI}} in {{Medical Imaging}}, and {{Ethical}} and
  {{Philosophical Issues}} in {{Medical Imaging}}}, 2023, vol. 14242, pp.
  163--171.

\bibitem{schwartz2021association}
C.~Schwartz \emph{et~al.}, ``Association of population screening for breast
  cancer risk with use of mammography among women in medically underserved
  racial and ethnic minority groups,'' \emph{JAMA Netw. Open}, vol.~4, no.~9,
  pp. e2\,123\,751--e2\,123\,751, 2021.

\bibitem{picarra_analysing_2023}
C.~Piçarra and B.~Glocker, ``Analysing race and sex bias in brain age
  prediction,'' in \emph{Clin. {{Image-Based Proced}}. {{Fairness AI Med}}.
  {{Imaging Ethical Philos}}. {{Issues Med}}. {{Imaging}}}, 2023, pp. 194--204.

\bibitem{du_unveiling_2023}
Y.~Du, Y.~Xue, R.~Dharmakumar, and S.~A. Tsaftaris, ``Unveiling fairness biases
  in deep learning-based brain {{MRI}} reconstruction,'' in \emph{Clin.
  {{Image-Based Proced}}. {{Fairness AI Med}}. {{Imaging Ethical Philos}}.
  {{Issues Med}}. {{Imaging}}}, 2023, pp. 102--111.

\bibitem{sadafi_study_2023}
A.~Sadafi, M.~Hehr, N.~Navab, and C.~Marr, ``A study of age and sex bias in
  multiple instance learning based classification of acute myeloid leukemia
  subtypes,'' in \emph{Clin. {{Image-Based Proced}}. {{Fairness AI Med}}.
  {{Imaging Ethical Philos}}. {{Issues Med}}. {{Imaging}}}, 2023, pp. 256--265.

\bibitem{ganz2021assessing}
M.~Ganz, S.~H. Holm, and A.~Feragen, ``Assessing bias in medical ai,'' in
  \emph{Workshop {{Interpret ML Heal}}. {{Int Connference Mach Learn ICML}}},
  2021.

\bibitem{groh2021evaluating}
M.~Groh \emph{et~al.}, ``Evaluating deep neural networks trained on clinical
  images in dermatology with the fitzpatrick 17k dataset,'' in \emph{Proc.
  {{IEEECVF Conf}}. {{Comput}}. {{Vis}}. {{Pattern Recognit}}.}, 2021, pp.
  1820--1828.

\bibitem{tian2024fairseg}
Y.~Tian \emph{et~al.}, ``{{FairSeg}}: A large-scale medical image segmentation
  dataset for fairness learning using segment anything model with fair
  error-bound scaling,'' in \emph{Int. {{Conf}}. {{Learn}}. {{Represent}}.
  {{ICLR}}}, 2024.

\bibitem{campos2014gender}
A.~Campos, ``Gender differences in imagery,'' \emph{Personal. Individ.
  Differ.}, vol.~59, pp. 107--111, 2014.

\bibitem{borgwardt2006integrating}
K.~M. Borgwardt \emph{et~al.}, ``Integrating structured biological data by
  kernel maximum mean discrepancy,'' \emph{Bioinformatics}, vol.~22, no.~14,
  pp. e49--e57, 2006.

\bibitem{barron2020generalization}
J.~T. Barron, ``A generalization of {{Otsu}}’s method and minimum error
  thresholding,'' in \emph{Comput. {{Vision}}–{{ECCV}} 2020 16th {{Eur}}.
  {{Conf}}. {{Glasg}}. {{UK August}} 23–28 2020 {{Proc}}. {{Part V}} 16},
  2020, pp. 455--470.

\bibitem{kearns2018preventing}
M.~Kearns, S.~Neel, A.~Roth, and Z.~S. Wu, ``Preventing fairness
  gerrymandering: {{Auditing}} and learning for subgroup fairness,'' in
  \emph{Int. {{Conf}}. {{Mach}}. {{Learn}}.}, 2018, pp. 2564--2572.

\bibitem{irvin2019chexpert}
J.~Irvin \emph{et~al.}, ``Chexpert: {{A}} large chest radiograph dataset with
  uncertainty labels and expert comparison,'' in \emph{Proc. {{AAAI Conf}}.
  {{Artif}}. {{Intell}}.}, vol.~33, no.~01, 2019, pp. 590--597.

\bibitem{wang2017chestx}
X.~Wang \emph{et~al.}, ``Chestx-ray8: {{Hospital-scale}} chest x-{{Ray}}
  database and benchmarks on weakly-supervised classification and localization
  of common thorax diseases,'' in \emph{Proc. {{IEEE Conf}}. {{Comput}}.
  {{Vis}}. {{Pattern Recognit}}.}, 2017, pp. 2097--2106.

\bibitem{johnson2019mimic}
A.~E. Johnson \emph{et~al.}, ``{{MIMIC-CXR}}, a de-{{Identified}} publicly
  available database of chest radiographs with free-text reports,'' \emph{Sci.
  Data}, vol.~6, no.~1, pp. 1--8, 2019.

\bibitem{bustos2020padchest}
A.~Bustos, A.~Pertusa, J.-M. Salinas, and M.~De~La Iglesia-Vaya, ``Padchest:
  {{A}} large chest x-{{Ray}} image dataset with multi-label annotated
  reports,'' \emph{Med. Image Anal.}, vol.~66, p. 101797, 2020.

\bibitem{borghesi2020covid}
A.~Borghesi and R.~Maroldi, ``{{COVID-19}} outbreak in {{Italy}}:
  {{Experimental}} chest {{X-ray}} scoring system for quantifying and
  monitoring disease progression,'' \emph{Radiol. Med. (Torino)}, vol. 125, pp.
  509--513, 2020.

\bibitem{BS-Net2021}
A.~Signoroni \emph{et~al.}, ``{{BS-Net}}: {{Learning COVID-19}} pneumonia
  severity on a large {{Chest X-Ray}} dataset,'' \emph{Med. Image Anal.}, p.
  102046, 2021.

\bibitem{shiraishi2000development}
J.~Shiraishi \emph{et~al.}, ``Development of a digital image database for chest
  radiographs with and without a lung nodule: {{Receiver}} operating
  characteristic analysis of radiologists' detection of pulmonary nodules,''
  \emph{Am. J. Roentgenol.}, vol. 174, no.~1, pp. 71--74, 2000.

\bibitem{cohen2020covidProspective}
\BIBentryALTinterwordspacing
J.~P. Cohen \emph{et~al.}, ``{COVID-19} image data collection: Prospective
  predictions are the future,'' \emph{CoRR}, vol. abs/2006.11988, 2020.
  [Online]. Available: \url{https://arxiv.org/abs/2006.11988}
\BIBentrySTDinterwordspacing

\bibitem{jaeger2014two}
S.~Jaeger \emph{et~al.}, ``Two public chest {{X-ray}} datasets for
  computer-aided screening of pulmonary diseases,'' \emph{Quant. Imaging Med.
  Surg.}, vol.~4, no.~6, p. 475, 2014.

\bibitem{kramer1993national}
B.~S. Kramer, J.~Gohagan, P.~C. Prorok, and C.~Smart, ``A {{National Cancer
  Institute}} sponsored screening trial for prostatic, lung, colorectal, and
  ovarian cancers,'' \emph{Cancer}, vol.~71, no.~S2, pp. 589--593, 1993.

\bibitem{nguyen2022vindr}
H.~Q. Nguyen \emph{et~al.}, ``{{VinDr-CXR}}: {{An}} open dataset of chest
  {{X-rays}} with radiologist’s annotations,'' \emph{Sci. Data}, vol.~9,
  no.~1, p. 429, 2022.

\bibitem{ettinger2017non}
D.~S. Ettinger \emph{et~al.}, ``Non–small cell lung cancer, version 5.2017,
  {{NCCN}} clinical practice guidelines in oncology,'' \emph{J. Natl. Compr.
  Canc. Netw.}, vol.~15, no.~4, pp. 504--535, 2017.

\bibitem{shakouri2021covid19}
S.~Shakouri \emph{et~al.}, ``{{COVID19-CT-dataset}}: {{An}} open-access chest
  {{CT}} image repository of 1000+ patients with confirmed {{COVID-19}}
  diagnosis,'' \emph{BMC Res. Notes}, vol.~14, no.~1, pp. 1--3, 2021.

\bibitem{national2011national}
N.~L. S. T.~R. Team \emph{et~al.}, ``The national lung screening trial:
  {{Overview}} and study design,'' \emph{Radiology}, vol. 258, no.~1, p. 243,
  2011.

\bibitem{pedrosa2019lndb}
\BIBentryALTinterwordspacing
J.~Pedrosa \emph{et~al.}, ``Lndb: {A} lung nodule database on computed
  tomography,'' \emph{CoRR}, vol. abs/1911.08434, 2019. [Online]. Available:
  \url{http://arxiv.org/abs/1911.08434}
\BIBentrySTDinterwordspacing

\bibitem{afshar2021covid}
P.~Afshar \emph{et~al.}, ``{{COVID-CT-MD}}, {{COVID-19}} computed tomography
  scan dataset applicable in machine learning and deep learning,'' \emph{Sci.
  Data}, vol.~8, no.~1, p. 121, 2021.

\bibitem{heller2019kits19}
\BIBentryALTinterwordspacing
N.~Heller \emph{et~al.}, ``The kits19 challenge data: 300 kidney tumor cases
  with clinical context, {CT} semantic segmentations, and surgical outcomes,''
  \emph{CoRR}, vol. abs/1904.00445, 2019. [Online]. Available:
  \url{http://arxiv.org/abs/1904.00445}
\BIBentrySTDinterwordspacing

\bibitem{bejarano2019longitudinal}
T.~Bejarano, M.~De~Ornelas-Couto, and I.~B. Mihaylov, ``Longitudinal fan-beam
  computed tomography dataset for head-and-neck squamous cell carcinoma
  patients,'' \emph{Med. Phys.}, vol.~46, no.~5, pp. 2526--2537, 2019.

\bibitem{cao2003image}
F.~Cao \emph{et~al.}, ``Image database for digital hand atlas,'' in \emph{Med
  {{Imaging}} 2003 {{PACS Integr Med Inf Syst Eval}}}, 2003.

\bibitem{yan2018deeplesion}
K.~Yan, X.~Wang, L.~Lu, and R.~M. Summers, ``{{DeepLesion}}: {{Automated}}
  mining of large-scale lesion annotations and universal lesion detection with
  deep learning,'' \emph{J. Med. Imaging}, vol.~5, no.~3, p. 036501, 2018.

\bibitem{age1999age}
A.-R. E. D. S.~R. Group \emph{et~al.}, ``The age-related eye disease study
  ({{AREDS}}): {{Design}} implications {{AREDS}} report no. 1,'' \emph{Control.
  Clin. Trials}, vol.~20, no.~6, p. 573, 1999.

\bibitem{kovalyk2022papila}
O.~Kovalyk \emph{et~al.}, ``{{PAPILA}}: {{Dataset}} with fundus images and
  clinical data of both eyes of the same patient for glaucoma assessment,''
  \emph{Sci. Data}, vol.~9, no.~1, p. 291, 2022.

\bibitem{diaz2019automatic}
M.~D{\'\i}az \emph{et~al.}, ``Automatic segmentation of the foveal avascular
  zone in ophthalmological oct-a images,'' \emph{PloS one}, vol.~14, no.~2, p.
  e0212364, 2019.

\bibitem{he2019retinal}
Y.~He \emph{et~al.}, ``Retinal layer parcellation of optical coherence
  tomography images: {{Data}} resource for multiple sclerosis and healthy
  controls,'' \emph{Data Brief}, vol.~22, pp. 601--604, 2019.

\bibitem{farsiu2014quantitative}
S.~Farsiu \emph{et~al.}, ``Quantitative classification of eyes with and without
  intermediate age-related macular degeneration using optical coherence
  tomography,'' \emph{Ophthalmology}, vol. 121, no.~1, pp. 162--172, 2014.

\bibitem{kwitt2013localizing}
R.~Kwitt, N.~Vasconcelos, S.~Razzaque, and S.~Aylward, ``Localizing target
  structures in ultrasound video–a phantom study,'' \emph{Med. Image Anal.},
  vol.~17, no.~7, pp. 712--722, 2013.

\bibitem{bernard2018deep}
O.~Bernard \emph{et~al.}, ``Deep learning techniques for automatic {{MRI}}
  cardiac multi-structures segmentation and diagnosis: {{Is}} the problem
  solved?'' \emph{IEEE Trans. Med. Imaging}, vol.~37, no.~11, pp. 2514--2525,
  2018.

\bibitem{petersen2015uk}
S.~E. Petersen \emph{et~al.}, ``{{UK Biobank}}’s cardiovascular magnetic
  resonance protocol,'' \emph{J. Cardiovasc. Magn. Reson.}, vol.~18, no.~1, pp.
  1--7, 2015.

\bibitem{adhd2012adhd}
A.-. {consortium}, ``The {{ADHD-200}} consortium: A model to advance the
  translational potential of neuroimaging in clinical neuroscience,''
  \emph{Front. Syst. Neurosci.}, vol.~6, p.~62, 2012.

\bibitem{marcus2007open}
D.~S. Marcus \emph{et~al.}, ``Open {{Access Series}} of {{Imaging Studies}}
  ({{OASIS}}): {{Cross-sectional MRI}} data in young, middle aged, nondemented,
  and demented older adults,'' \emph{J. Cogn. Neurosci.}, vol.~19, no.~9, pp.
  1498--1507, 2007.

\bibitem{di2014autism}
A.~Di~Martino \emph{et~al.}, ``The autism brain imaging data exchange:
  {{Towards}} a large-scale evaluation of the intrinsic brain architecture in
  autism,'' \emph{Mol. Psychiatry}, vol.~19, no.~6, pp. 659--667, 2014.

\bibitem{marek2011parkinson}
K.~Marek \emph{et~al.}, ``The {{Parkinson}} progression marker initiative
  ({{PPMI}}),'' \emph{Prog. Neurobiol.}, vol.~95, no.~4, pp. 629--635, 2011.

\bibitem{menze2014multimodal}
B.~H. Menze \emph{et~al.}, ``The multimodal brain tumor image segmentation
  benchmark ({{BRATS}}),'' \emph{IEEE Trans. Med. Imaging}, vol.~34, no.~10,
  pp. 1993--2024, 2014.

\bibitem{frazier2008diagnostic}
J.~A. Frazier \emph{et~al.}, ``Diagnostic and sex effects on limbic volumes in
  early-onset bipolar disorder and schizophrenia,'' \emph{Schizophr. Bull.},
  vol.~34, no.~1, pp. 37--46, 2008.

\bibitem{shafto2014cambridge}
M.~A. Shafto \emph{et~al.}, ``The {{Cambridge Centre}} for {{Ageing}} and
  {{Neuroscience}} ({{Cam-CAN}}) study protocol: A cross-sectional, lifespan,
  multidisciplinary examination of healthy cognitive ageing,'' \emph{BMC
  Neurol.}, vol.~14, no.~1, pp. 1--25, 2014.

\bibitem{milham2011international}
M.~Milham, M.~Mennes, D.~Gutman \emph{et~al.}, ``The international neuroimaging
  data-sharing initiative ({{INDI}}) and the functional connectomes project,''
  in \emph{17th {{Annu}}. {{Meet}}. {{Organ}}. {{Hum}}. {{Brain Mapp}}.}, 2011.

\bibitem{jack2008alzheimer}
C.~R. Jack~Jr \emph{et~al.}, ``The {{Alzheimer}}'s disease neuroimaging
  initiative ({{ADNI}}): {{MRI}} methods,'' \emph{J Magn Reson Imaging J Int
  Soc Magn Reson Med}, vol.~27, no.~4, pp. 685--691, 2008.

\bibitem{karcher2021abcd}
N.~R. Karcher and D.~M. Barch, ``The {{ABCD}} study: {{Understanding}} the
  development of risk for mental and physical health outcomes,''
  \emph{Neuropsychopharmacology}, vol.~46, no.~1, pp. 131--142, 2021.

\bibitem{codella2019skin}
N.~C.~F. Codella \emph{et~al.}, ``Skin lesion analysis toward melanoma
  detection: {A} challenge at the 2017 international symposium on biomedical
  imaging (isbi), hosted by the international skin imaging collaboration
  {(ISIC)},'' in \emph{2018 {{IEEE}} 15th {{Int}}. {{Symp}}. {{Biomed}}.
  {{Imaging ISBI}}}.\hskip 1em plus 0.5em minus 0.4em\relax {IEEE}, 2018, pp.
  168--172.

\bibitem{rotemberg2021patient}
V.~Rotemberg \emph{et~al.}, ``A patient-centric dataset of images and metadata
  for identifying melanomas using clinical context,'' \emph{Sci. Data}, vol.~8,
  no.~1, p.~34, 2021.

\bibitem{groh2022towards}
\BIBentryALTinterwordspacing
M.~Groh, C.~Harris, R.~Daneshjou, O.~Badri, and A.~Koochek, ``Towards
  transparency in dermatology image datasets with skin tone annotations by
  experts, crowds, and an algorithm,'' \emph{Proc. {ACM} Hum. Comput.
  Interact.}, vol.~6, no. {CSCW2}, pp. 1--26, 2022. [Online]. Available:
  \url{https://doi.org/10.1145/3555634}
\BIBentrySTDinterwordspacing

\bibitem{ganin2015unsupervised}
Y.~Ganin and V.~Lempitsky, ``Unsupervised domain adaptation by
  backpropagation,'' in \emph{Int. {{Conf}}. {{Mach}}. {{Learn}}.}, 2015, pp.
  1180--1189.

\bibitem{creager2019flexibly}
E.~Creager \emph{et~al.}, ``Flexibly fair representation learning by
  disentanglement,'' in \emph{Int. {{Conf}}. {{Mach}}. {{Learn}}.}, 2019, pp.
  1436--1445.

\bibitem{chuang2020debiased}
C.-Y. Chuang, J.~Robinson, Y.-C. Lin, A.~Torralba, and S.~Jegelka, ``Debiased
  contrastive learning,'' \emph{Adv. Neural Inf. Process. Syst.}, vol.~33, pp.
  8765--8775, 2020.

\bibitem{jungo}
A.~Jungo and M.~Reyes, ``Assessing reliability and challenges of uncertainty
  estimations for medical image segmentation,'' in \emph{Med. {{Image Comput}}.
  {{Comput}}. {{Assist}}. {{Interv}}. – {{MICCAI}} 2019}, 2019, pp. 48--56.

\bibitem{ricci_lara_addressing_2022}
M.~A. Ricci~Lara, R.~Echeveste, and E.~Ferrante, ``Addressing fairness in
  artificial intelligence for medical imaging.'' \emph{Nat. Commun.}, vol.~13,
  no.~1, p. 4581, 2022-08.

\bibitem{clark2013cancer}
K.~Clark \emph{et~al.}, ``The {{Cancer Imaging Archive}} ({{TCIA}}):
  {{Maintaining}} and operating a public information repository,'' \emph{J.
  Digit. Imaging}, vol.~26, no.~6, pp. 1045--1057, 2013.

\bibitem{webb2022addressing}
E.~K. Webb, J.~A. Etter, and J.~A. Kwasa, ``Addressing racial and phenotypic
  bias in human neuroscience methods,'' \emph{Nat. Neurosci.}, vol.~25, no.~4,
  pp. 410--414, 2022.

\bibitem{gonzalez2021disparities}
D.~Gonzalez, ``Disparities in {{COVID-19}} rates among various demographics and
  lack of racial representation in medical texts,'' \emph{Celebr. Learn.},
  2021.

\bibitem{wang2021directional}
A.~Wang and O.~Russakovsky, ``Directional bias amplification,'' in \emph{Int.
  {{Conf}}. {{Mach}}. {{Learn}}.}, 2021, pp. 10\,882--10\,893.

\bibitem{karkkainen2021fairface}
K.~Karkkainen and J.~Joo, ``Fairface: {{Face}} attribute dataset for balanced
  race, gender, and age for bias measurement and mitigation,'' in \emph{Proc.
  {{IEEECVF Winter Conf}}. {{Appl}}. {{Comput}}. {{Vis}}.}, 2021, pp.
  1548--1558.

\bibitem{liu2015deep}
Z.~Liu, P.~Luo, X.~Wang, and X.~Tang, ``Deep learning face attributes in the
  wild,'' in \emph{Proc. {{IEEE Int}}. {{Conf}}. {{Comput}}. {{Vis}}.}, 2015,
  pp. 3730--3738.

\bibitem{mccradden2020ethical}
M.~D. McCradden, S.~Joshi, M.~Mazwi, and J.~A. Anderson, ``Ethical limitations
  of algorithmic fairness solutions in health care machine learning,''
  \emph{Lancet Digit. Health}, vol.~2, no.~5, pp. e221--e223, 2020.

\bibitem{fairbairn2020sex}
T.~A. Fairbairn \emph{et~al.}, ``Sex differences in coronary computed
  tomography angiography–derived fractional flow reserve: {{Lessons}} from
  {{ADVANCE}},'' \emph{Cardiovasc. Imaging}, vol.~13, no.~12, pp. 2576--2587,
  2020.

\bibitem{lee2020biobert}
J.~Lee \emph{et~al.}, ``{{BioBERT}}: A pre-trained biomedical language
  representation model for biomedical text mining,'' \emph{Bioinformatics},
  vol.~36, no.~4, pp. 1234--1240, 2020.

\bibitem{MedicalGPT}
\BIBentryALTinterwordspacing
M.~Xu, ``{{MedicalGPT}}: {{Training}} medical {{GPT}} model,'' 2023. [Online].
  Available: \url{https://github.com/shibing624/MedicalGPT}
\BIBentrySTDinterwordspacing

\bibitem{radford2021learning}
A.~Radford \emph{et~al.}, ``Learning transferable visual models from natural
  language supervision,'' in \emph{Int. {{Conf}}. {{Mach}}. {{Learn}}.}, 2021,
  pp. 8748--8763.

\bibitem{eslami2023pubmedclip}
S.~Eslami, C.~Meinel, and G.~De~Melo, ``Pubmedclip: {{How}} much does clip
  benefit visual question answering in the medical domain?'' in \emph{Find.
  {{Assoc}}. {{Comput}}. {{Linguist}}. {{EACL}} 2023}, 2023, pp. 1181--1193.

\bibitem{lin2023pmc}
W.~Lin \emph{et~al.}, ``Pmc-clip: {{Contrastive}} language-image pre-training
  using biomedical documents,'' in \emph{Int. {{Conf}}. {{Med}}. {{Image
  Comput}}. {{Comput}}.-{{Assist}}. {{Interv}}.}, 2023, pp. 525--536.

\bibitem{lai2024carzero}
\BIBentryALTinterwordspacing
H.~Lai \emph{et~al.}, ``Carzero: Cross-attention alignment for radiology
  zero-shot classification,'' \emph{CoRR}, vol. abs/2402.17417, 2024. [Online].
  Available: \url{https://doi.org/10.48550/arXiv.2402.17417}
\BIBentrySTDinterwordspacing

\bibitem{kirillov2023segment}
A.~Kirillov \emph{et~al.}, ``Segment anything,'' in \emph{{IEEE/CVF} Conf.
  Comput. Vis.}\hskip 1em plus 0.5em minus 0.4em\relax {IEEE}, 2023, pp.
  3992--4003.

\bibitem{MedSAM}
J.~Ma \emph{et~al.}, ``Segment anything in medical images,'' \emph{Nat.
  Commun.}, vol.~15, no.~1, p. 654, 2024.

\bibitem{quan2024slide}
Q.~Quan, F.~Tang, Z.~Xu, H.~Zhu, and S.~K. Zhou, ``Slide-{{SAM}}: {{Medical
  SAM}} meets sliding window,'' in \emph{Med. {{Imaging Deep Learn}}.}, 2024.

\bibitem{wang2023sammed3d}
\BIBentryALTinterwordspacing
H.~Wang \emph{et~al.}, ``Sam-med3d,'' 2023. [Online]. Available:
  \url{https://doi.org/10.48550/arXiv.2310.15161}
\BIBentrySTDinterwordspacing

\bibitem{li2023survey}
\BIBentryALTinterwordspacing
Y.~Li, M.~Du, R.~Song, X.~Wang, and Y.~Wang, ``A survey on fairness in large
  language models,'' \emph{CoRR}, vol. abs/2308.10149, 2023. [Online].
  Available: \url{https://doi.org/10.48550/arXiv.2308.10149}
\BIBentrySTDinterwordspacing

\bibitem{Wang2021MitigateGenderBiasInImageSearch}
J.~Wang, Y.~Liu, and X.~E. Wang, ``Are gender-neutral queries really
  gender-neutral? {{Mitigating}} gender bias in image search,'' in
  \emph{{{EMNLP}}}, 2021.

\bibitem{berg2022prompt}
H.~Berg \emph{et~al.}, ``A prompt array keeps the bias away: {{Debiasing}}
  vision-language models with adversarial learning,'' in \emph{Proc. 2nd
  {{Conf}}. {{Asia-Pac}}. {{Chapter Assoc}}. {{Comput}}. {{Linguist}}. 12th
  {{Int}}. {{Jt}}. {{Conf}}. {{Nat}}. {{Lang}}. {{Process}}. {{Vol}}. 1 {{Long
  Pap}}.}, 2022-11, pp. 806--822.

\bibitem{hendricks2018women}
L.~A. Hendricks, K.~Burns, K.~Saenko, T.~Darrell, and A.~Rohrbach, ``Women also
  snowboard: {{Overcoming}} bias in captioning models,'' in \emph{Proc.
  {{Eur}}. {{Conf}}. {{Comput}}. {{Vis}}. {{ECCV}}}, 2018, pp. 771--787.

\bibitem{xu2024apple}
\BIBentryALTinterwordspacing
Z.~Xu, F.~Tang, Q.~Quan, Q.~Yao, and S.~K. Zhou, ``{APPLE:} adversarial
  privacy-aware perturbations on latent embedding for unfairness mitigation,''
  \emph{CoRR}, vol. abs/2403.05114, 2024. [Online]. Available:
  \url{https://doi.org/10.48550/arXiv.2403.05114}
\BIBentrySTDinterwordspacing

\bibitem{jin2024universal}
\BIBentryALTinterwordspacing
R.~Jin, W.~Deng, M.~Chen, and X.~Li, ``Universal debiased editing on foundation
  models for fair medical image classification,'' \emph{CoRR}, vol.
  abs/2403.06104, 2024. [Online]. Available:
  \url{https://doi.org/10.48550/arXiv.2403.06104}
\BIBentrySTDinterwordspacing

\bibitem{collins2021protocol}
G.~S. Collins \emph{et~al.}, ``Protocol for development of a reporting
  guideline ({{TRIPOD-AI}}) and risk of bias tool ({{PROBAST-AI}}) for
  diagnostic and prognostic prediction model studies based on artificial
  intelligence,'' \emph{BMJ open}, vol.~11, no.~7, p. e048008, 2021.

\bibitem{takkouche2011prisma}
B.~Takkouche and G.~Norman, ``{{PRISMA}} statement,'' \emph{Epidemiology},
  vol.~22, no.~1, p. 128, 2011.

\end{thebibliography}
